# Intelligent Identification and Repair of Design Defects in BIM via Domain-Specific Large Language Models

Jia-Rui Lin [a,b], Yun-Hong Cai [a,b], Xiang-Rui Ni [a,b,c], Peng Pan [a,b,*]

[a] Department of Civil Engineering, Tsinghua University, Beijing, China, 100084

[b] Key Laboratory of Digital Construction and Digital Twin, Ministry of Housing and Urban-Rural Development, Beijing, China, 100084

[c] PetroChina Planning & Engineering Institute, Beijing, China, 100083

## Abstract

Existing methods lack a generalized approach to efficiently identify and resolve the diversity of design defects in BIM. Therefore, this study proposes an integrated framework to identify and repair various defects in BIM via domain-specific LLMs. Firstly, a BIM-to-Text method with component-balanced chunking is introduced to bridge BIM data with LLMs. Then, prompt learning with rule injection, few-shot prompting and RAG is proposed to identify defects and generate repair suggestions. Meanwhile, a hallucination control strategy combining key identifier validation and token-length thresholds is introduced to ensure reliability. Experiments show capability expansion yields 85% identification accuracy versus 70% for traditional rule checking, achieving a 94% rate of reasonable repair suggestions. Moreover, the proposed hallucination control further increased accuracy from 64% to 85%, eliminating 92.5% of hallucinations in a single intervention round. This study establishes an end-to-end prototype from raw BIM data input, through defect identification, to repair suggestion generation.

## 1 Introduction

To address the increasing complexity of modern infrastructure and ensure information consistency, Building Information Modeling (BIM) technology has been widely adopted in construction engineering [1]. This technology enables integrated management of multi-source data across the design, construction, and operation stages, facilitating efficient stakeholder collaboration and enhancing overall project quality [2]. Among these stages, design quality is of paramount importance, as it fundamentally determines the efficiency and reliability of the entire project lifecycle [3].

Nevertheless, in practical applications, BIM data produced during the design phase are often affected by human factors, leading to various quality issues such as missing information, non-compliant parameters, and inaccurate representations of spatial relationships [4]. This study collectively refers to such quality problems induced by human factors as design defects. If these defects are not identified and addressed prior to design delivery, they may impose severe risks on project schedule, cost, and quality during subsequent construction and operation stages [5]. Accordingly, establishing reliable mechanisms for defect identification and resolution at the design stage is not only critical for ensuring smooth project implementation within the prescribed constraints of time, cost, and quality, but also essential for effective risk control throughout the entire project lifecycle.

In engineering practice, defect identification remains predominantly dependent on manual inspection, while the existing automated methods are still limited. Current studies on defect identification primarily adopt rule-based approaches and collision-detection approaches, both of which present inherent limitations. Rule-based approaches leverage techniques such as natural language processing [6], knowledge graphs [7], and deep learning [8, 9] to interpret regulatory texts into computable rules, followed by model auditing based on rule checking. Collision-detection approaches, on the other hand, are designed to address geometric and spatial conflicts within models. However, these two types of approaches often target a certain category of defects and cannot be directly transferred across different scenarios. Since BIM quality issues are diverse and interrelated, and they often involve complex semantic contexts beyond explicit numerical violations, purely rule-based approaches struggle to capture implicit design conventions. Consequently, they require separate interpretation, coding, and maintenance of rules for each defect type, resulting in insufficient generalizability. Meanwhile, collision detection lacks semantic support, leading to false alarms and the omission of semantic defects related to functionality, materials, or fire resistance ratings [10]. Therefore, in the defect identification stage, reliance solely on manual inspection or rule-based methods is insufficient, and tailored methods must often be developed for different defect categories.

Beyond identification, defect repair still largely depends on manual processes. At present, repairing such design defects typically requires interdisciplinary design teams to discuss and devise corrective solutions according to various building codes and their rich experiences, a process that is both time-intensive and resource-demanding. Because effective repair requires synthesizing rigid regulatory constraints with complex engineering expertise, automating this reasoning process remains a fundamental challenge. This situation highlights the urgent need for automated repair methods. However, research addressing BIM defect repair remains scarce and is mostly confined to localized scenarios. Hence, for engineering applications, there is a pressing demand to develop automated repair methods that can adapt across defect categories, are supported by regulatory standards, and can improve the quality of design outputs while addressing the current research gap.

Large Language Models (LLMs) represent a recent breakthrough in the field of artificial intelligence and provide new opportunities for addressing BIM defect identification and repair. Representative models such as ChatGPT rely on large-scale unsupervised pre-training on massive text corpora, enabling them to acquire rich semantic and world knowledge. As a result, they possess strong natural language understanding capabilities and can generate human-like responses in natural language [11]. Numerous studies have demonstrated that LLMs possess exceptional transfer learning capabilities, enabling them to be fine-tuned with a limited amount of domain-specific data and further enhanced through methods such as prompt learning and retrieval-augmented generation (RAG). These techniques strengthen the model's understanding of domain knowledge and its adaptability to specific tasks, thereby allowing it to efficiently address domain-specific problems and achieve outstanding performance [12]. Consequently, domain-specific LLMs offer new possibilities for the unified identification and repair of various design defects. However, LLMs are prone to hallucination during the generation process, where the model may produce content that is factually incorrect or semantically irrelevant to the input. This issue severely restricts their application in engineering design scenarios, where the reliability of results is of critical importance.

In summary, the automated identification and repair of multiple types of BIM design defects still lack a unified and generalizable method. While LLMs offer the potential for more advanced and efficient solutions, their effective application in engineering relies on domain-specific adaptation. Consequently, to address this specific task, combining domain fine-tuning with RAG and prompt learning is deduced to achieve a better balance between accuracy and generalization than any single specialized method. However, given the high precision and reliability required in engineering applications, the potential hallucination issue during the generation process must be rigorously controlled [13].

This study aims to develop a domain-specific LLM-based integrated framework for the identification and repair of multiple types of BIM design defects, addressing the practical engineering demand for intelligent and efficient solutions, and ultimately establishing a reliable end-to-end prototype. Furthermore, this research reveals the respective applicability boundaries and synergistic effects of LLM fine-tuning, RAG, and prompt learning in highly specialized engineering tasks. The remainder of this paper is organized as follows. Section 2 reviews the relevant research on BIM design defect identification and repair, and analyzes the current applications and limitations of LLMs in engineering design. Section 3 first introduces a BIM-to-Text method with component-balanced chunking to bridge BIM models with LLMs. Then, prompt templates combining rule injection and few-shot prompting are designed to achieve unified identification of multiple defect types. On this basis, a RAG-based method is developed to generate repair suggestions for detected defects, followed by hallucination control strategies that combine key identifier validation and token-length threshold to mitigate potential hallucinated outputs. Finally, the experimental setup to validate the proposed methods is presented. Section 4 presents the experimental results and provides a comprehensive discussion. Section 5 concludes the paper and outlines directions for future research.

## 2 Literature Review

### 2.1 BIM Design Defects

BIM design defects exert significant negative impacts on project progress, cost, and quality [14]. Studies have shown that such defects frequently lead to rework, which in turn causes schedule delays, cost overruns, and may intensify disputes and conflicts among stakeholders [15]. Given their inevitability and potentially severe consequences, design defects are regarded as one of the primary sources of risk and performance decline in construction projects [16]. Accordingly, the identification and prevention of design defects are essential for ensuring project progress and represent one of the core issues that must be addressed in BIM applications.

Various classification methods and perspectives have been proposed in existing research on BIM design defects. Lopez et al. [17], from the perspective of human error, categorized design defects into three groups: knowledge errors, operational errors, and non-compliance errors. Other studies, focusing on the manifestation of defects, have classified them into several common types, including inaccuracies or missing information in design results (e.g., incomplete specifications or missing components), computational errors or mistakes in dimensional annotations, conflicts or collisions between components, and violations of regulatory provisions in design outcomes [10]. Some scholars have further distinguished between active errors and passive omissions, with the former referring to deviations introduced by designers during operation, and the latter referring to necessary content being inadvertently omitted from the design output [18].

Building on the above research, this study categorizes design defects into three groups with a focus on BIM model data, namely integrity defects, rationality defects, and compliance defects. Integrity defects refer to missing components or insufficient attribute information of components in the model. Such defects typically arise when specific application scenarios require certain model data that are not provided. Rationality defects refer to cases where the dimensions, attribute information, or relative spatial relationships of components deviate from design common sense or engineering experience, thereby lacking rationality. These defects are particularly complex in architectural design, as design solutions are often difficult to evaluate strictly in terms of right or wrong; although they may not explicitly violate codes or standards, they can still deviate from conventional design practices. Compliance defects refer to cases where model components fail to meet the requirements of relevant design standards and regulations. These defects are the most common type in BIM models and are a major focus of design review in engineering practice.

**2.2 Defect Identification**

Research on automated design defect identification can be traced back to the 1960s, when Fenves first used decision tables to represent design rules and perform defect checking [19]. Since the beginning of the 21st century, BIM technology has provided more comprehensive capabilities for design [1]. At present the overall process of automated defect identification can be conceptually divided into two distinct stages identifying the rules that serve as the checking basis and identifying the target objects to be examined.

For identifying the rules that serve as the checking basis, including both rule interpretation and rule-algorithm-based defect detection, rule-based approaches are the primary focus of existing studies. These methods are relatively easy to understand and broadly applicable, but they still require extensive manual work to write query languages, pseudo-code, or annotated documents, resulting in low levels of automation [20]. For example, Nawari et al. [21] developed rule engines to verify the compliance of building models, and other researchers have constructed knowledge bases to store mechanical, electrical, and plumbing (MEP) design rules for BIM-based automated checking [22, 23]. This approach is beneficial in formalizing domain knowledge to identify defects that explicitly violate regulations; however, its applicability is limited, and the workload and maintenance challenges escalate sharply as data requirements expand [24]. In recent years, some studies have applied natural language processing [25] and ontology-based methods to achieve automatic or semi-automatic extraction of rules [26]. The extracted rules can then be transformed into executable formats such as IF-THEN logic or Semantic Web Rule Language (SWRL) through semantic modeling methods [27], or used to construct rule-based knowledge bases for BIM model checking [23]. At present, several mature software systems are available to perform automated compliance checking using rules expressed in formalized languages, such as the widely adopted Solibri Model Checker. However, these software systems still require manual encoding of rules, which entails high time and labor costs.

To identify the target objects to be examined some studies have further explored machine learning based methods. These methods aim to learn the mapping relationships between component features and design attributes, thereby enabling the automatic detection of abnormal or inconsistent elements within BIM models [28]. For example, some studies have adopted statistical analysis and clustering techniques to examine the parameter distributions of similar components in a model, define reasonable value ranges, and identify components that deviate significantly from these ranges [29]. While such methods are effective for identifying outliers in selected geometric parameters, their applicability to the more complex and semantically rich information in BIM models remains limited [30]. In addition, machine learning-based approaches often require large training datasets, iterative feature optimization, and high computational costs, which hinder their practical use in large-scale engineering projects [31].

In summary, existing defect identification methods remain highly dependent on extensive manual hard coding and have limited applicability. Although emerging technologies have been introduced into the field of design defect checking, their practical utility is still restricted.

### 2.3 Defect Repair for BIM

Research on BIM design defect repair remains relatively limited. Among existing studies, semantic augmentation methods are the most representative. These methods were initially developed to enrich models with additional semantic information in order to satisfy the requirements of specific scenarios or domains. Within this category, both rule-based and machine learning-based approaches have been applied to infer and supplement missing model data. Rule-based approaches construct expert systems by hard-coding the conditions that components must satisfy, and subsequently use these conditions to complete missing information. However, due to the need for manually encoding a large number of rules and the difficulty of addressing complex geometries, their applicability is constrained [32]. To overcome these limitations, recent studies have attempted to develop generative frameworks that automatically formulate code-compliant design modifications to resolve identified errors [33]. Despite these emerging efforts, the automation of defect repair and repair suggestion generation remains largely unexplored in current research [34].

As discussed in **Section 2.1**, this study focuses on the identification and repair of three categories of design defects: integrity, rationality, and compliance. **Table 1** summarizes the existing automated identification and repair techniques for these defect types, along with their limitations. For integrity defects, identification typically relies on integrity-rule-based existence checks, which can rapidly detect missing information or components under predefined conditions. Corresponding repair methods mainly adopt semantic augmentation, with rule-based expert systems and machine learning-based inference models being used to fill in missing information. Rationality defect identification often depends on statistical or clustering methods to detect parameter anomalies, but their coverage is limited, and extensive manual review is still required; similarly, repair is largely dependent on human expertise. Compliance defect identification relies on automated rule-checking based on regulatory codes and standards, in which provisions are translated into computer-executable logic for automatic comparison [8, 35]. However, rule extraction and encoding still require substantial manual effort, and repair continues to rely primarily on manual adjustment by designers.

**Table 1** BIM Design Defect Identification and Repair-Related Techniques

| Defect Type | Existing Identification Techniques | Limitations | Existing Repair Techniques | Limitations |
|---|---|---|---|---|
| Integrity | Integrity-rule-based checks [36] | Large coding workload | Rule-based methods; Machine learning [32] | Reliance on expertise, limited applicability; data acquisition is difficult, high cost |
| Rationality | Anomaly detection [29] | Limited coverage | - | - |
| Compliance | Regulation-based rules [21] | Complex code interpretation and encoding | - | - |

Taken together, the main challenges in this field can be summarized as follows:

(1) Different types of defects require different identification and repair methods, and a unified methodological and technical framework is lacking.

(2) Rule-based and collision-detection-based approaches require substantial manual hard coding, suffer from limited applicability, are incapable of addressing rationality-related issues, and deliver low identification accuracy.

(3) Research on defect repair methods remains scarce; the few existing methods are narrowly applicable, and in practical engineering, defect repair continues to depend heavily on manual analysis and decision-making.

**2.4 LLM for BIM-based Design**

LLMs have rapidly become a prominent development in natural language processing owing to their strong capabilities in semantic understanding, contextual reasoning, and coherent text generation. Through large-scale pretraining on extensive corpora, these models are able to extract high-level linguistic patterns and perform a wide range of language-intensive tasks with notable generalization performance. As model scales grow, LLMs such as GPT-5, Gemini, GLM-4, and Qwen have demonstrated increasingly sophisticated abilities, driving their adoption in applications that require complex knowledge interpretation and multi-step reasoning.

However, a major limitation that persists is the hallucination problem [37]. Fundamentally, hallucination is an inherent consequence of the autoregressive learning paradigm and architectural properties of transformer-based LLMs. Compounded by uneven pretraining data quality and the absence of truthfulness validation mechanisms, LLMs may generate false or unverifiable outputs [38], constraining their applicability in engineering and other safety-critical industries [39]. Existing studies have attempted to mitigate hallucinations by leveraging knowledge graphs [40] or augmenting model inputs with external information via RAG [41, 42]. To improve reliability and adapt LLMs to specific domains, researchers have developed domain-specialized models [43]. Methods for constructing domain-specific LLMs can be divided into parameter-tuning approaches, such as Low-Rank Adaptation (LoRA), and parameter-free approaches, which improve generation quality by optimizing or enriching model inputs, including prompt learning and RAG. Since our other study [44] has already implemented LoRA-based fine-tuning for domain adaptation, this study focuses on the parameter-free approaches of prompt learning and RAG for defect identification and repair tasks.

Prompt learning improves model output quality without altering parameters by optimizing input formats or adding semantic cues to guide generation. Its effectiveness mainly depends on task-specific prompt design [45]. Jin et al. [46] showed that such tailored prompts can enhance reliability, while recent strategies like self-consistency [47] and the Tree of Thought (ToT) method [48] further strengthen reasoning performance. Complementary to prompt learning, RAG enhances generation by retrieving domain-relevant knowledge from external sources and integrating it with user prompts [49]. As illustrated in **Fig. 1**, retrieved content is reranked to prioritize key information before input, effectively reducing hallucinations and improving accuracy. With well-designed prompts, the model can distinguish between general and domain-specific knowledge, ensuring reliable outputs [50]. Since this knowledge remains external, RAG provides a flexible, non-invasive enhancement for domain applications.

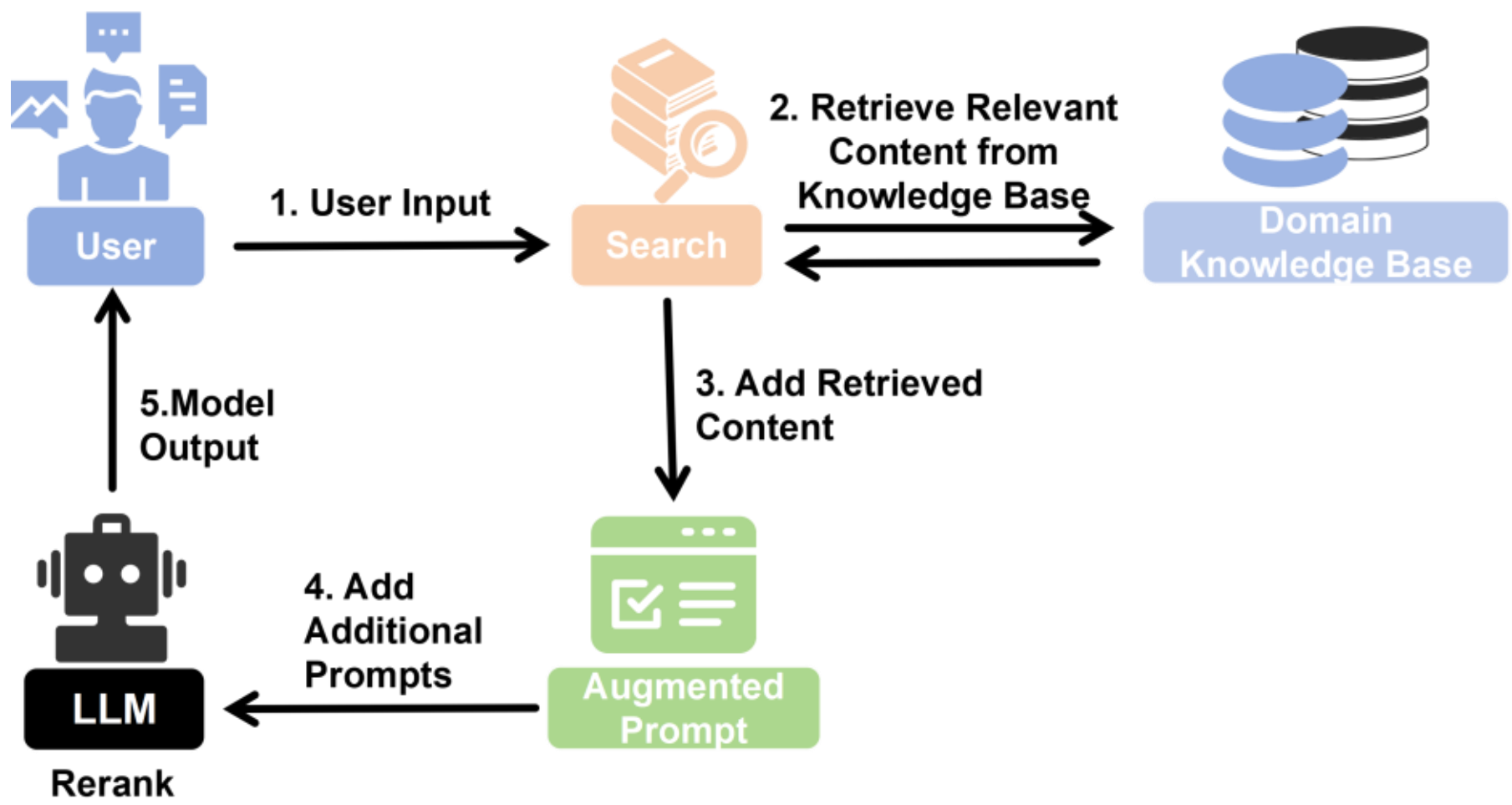

**Fig. 1.** Basic Workflow for RAG

Within civil engineering, researchers have begun exploring LLMs to improve project efficiency. However, most design-stage applications remain focused on auxiliary or detail-oriented tasks rather than core design processes [51]. Typical examples include inferring building types from BIM model names [52], extracting and structuring BIM information [53], and modifying Industry Foundation Classes (IFC)-formatted component information [54]. To advance beyond these auxiliary functions, recent cutting-edge studies have begun deploying LLM-based multi-agent frameworks to directly generate building models [55] and integrating AI with human expertise for customized design detailing [56]. Despite these significant advancements in generative design, in BIM design defect identification and repair, current LLM applications are still preliminary—mainly parsing regulatory texts and generating syntax trees without direct participation in the identification process, and most rely only on prompt learning without domain-specific optimization [57]. Existing studies also report issues such as the lack of rationality evaluation methods [54], data interoperability challenges, high infrastructure costs, and limited trust in model outputs [58].

Overall, existing research on BIM design defects has developed rule-based, statistical, and ontology-driven methods, yet these approaches remain fragmented and still rely heavily on manual encoding and expert judgment. Consequently, they struggle to capture the semantic complexity and contextual relationships involved in integrity, rationality, and compliance defects. The semantic understanding and reasoning capabilities of LLMs suggest that they could provide a more unified way to interpret heterogeneous information and support more adaptive defect analysis. However, current explorations of LLMs in BIM-related workflows remain preliminary and are mostly limited to auxiliary tasks rather than addressing defect identification or repair in a comprehensive manner. Systematic approaches for applying or constraining LLMs to support both stages of defect handling have not yet been established. This situation underscores the need for methods that can draw on the strengths of LLMs while maintaining sufficient control over their outputs, thereby enabling more reliable, scalable, and domain-aware automation for defect identification and repair.

## 3 Methodology

A related study [44] constructed a benchmark and BIM-derived dataset for evaluating and fine-tuning LLMs in BIM-based design, and developed the domain-specific model Qwen-BIM using Qwen2.5-14B-Instruct. The study showed that Qwen-BIM achieves a 21.0% average increase in G-Eval score compared to the base model and performs comparably to general LLMs with 671 billion parameters, providing foundational resources for this research. However, effectively leveraging these foundational capabilities for practical engineering tasks requires a specialized method to ensure reliability in defect identification and repair. Consequently, this chapter proposes an integrated framework for multi-category defect identification and repair suggestion generation in BIM design, with the overall architecture illustrated in **Fig. 2**.

The framework first introduces a BIM-to-Text process with component-balanced chunking, which converts BIM component information into semi-structured natural language to enhance model comprehension and efficiency. Building upon this data transformation, prompt templates that integrate rule injection and few-shot examples are utilized to guide the domain-specific LLM in identifying integrity, rationality, and compliance defects. For defect repair, a RAG method is employed to incorporate domain knowledge and produce well-grounded repair suggestions. In parallel, a hallucination control strategy that combines key-identifier validation with a token-length threshold is applied to enhance output accuracy and stability. Finally, as depicted in the upper validation module of the framework, the proposed methods are systematically validated through experiments conducted in a residential building BIM scenario.

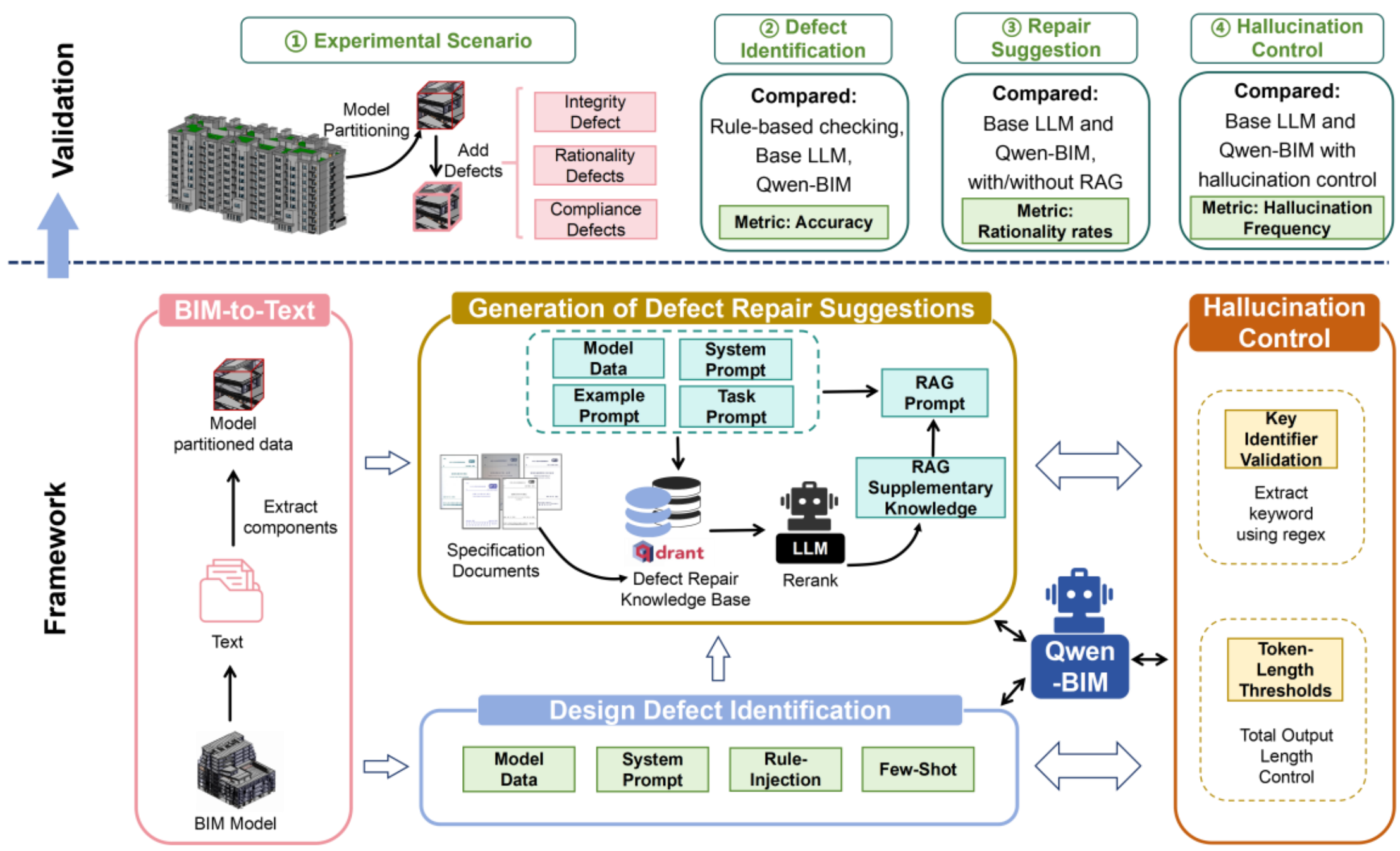

**Fig. 2.** Integrated framework for BIM design defect identification and repair suggestion generation

### 3.1 BIM-to-Text Method

To enable efficient processing of BIM data by LLMs, this study proposes a BIM-to-Text method with component-balanced chunking, which converts BIM component information into semi-structured natural language. The method consists of two main steps.

First, since the structured format of BIM data cannot be effectively utilized by LLMs in its original form, a semi-structured BIM-to-Text method is developed to convert BIM information into a more readable representation. This approach preserves key structural information by organizing data into a strict logical hierarchy, while introducing natural language expression logic through cohesive text templates, thereby enhancing the model's comprehension capability. The textual structure can be expressed as:

$$\text{Header Section} + n\,(\text{Categories}\times\text{Component Header Sections}) + n\,(\text{Components}\times\text{Component Data Sections}) \tag{1}$$

The detailed section patterns are as follows:

(1) Header Section: *This is part of a building model that contains several components such as [shear walls, structural beams, slabs, ...].*

(2) Component Header Section: *The following provides a description of the [component category], indicating its [spatial position] through coordinates and listing its [main parameters].*

(3) Component Data Section: *The [attribute$_1$] of the first [component category] is [value$_1$], ..., and the [attributen] of the n-th [component category] is [value$_n$].*

Specifically, the exact choices of [spatial position] and [main parameters] are tailored to balance computational efficiency and semantic richness. For [spatial position], start/end coordinates and reference elevations are utilized (e.g., for walls and beams) to encode 3D spatial boundaries and relative positions into a token-efficient text format, preventing the context-window explosion caused by processing raw geometric meshes. Regarding [main parameters], rather than extracting all available BIM properties, these parameters are selectively determined based on commonly used engineering attributes and the primary considerations of major building codes and standards. **Fig. 3** presents a concrete example of this method applied to an actual BIM model.

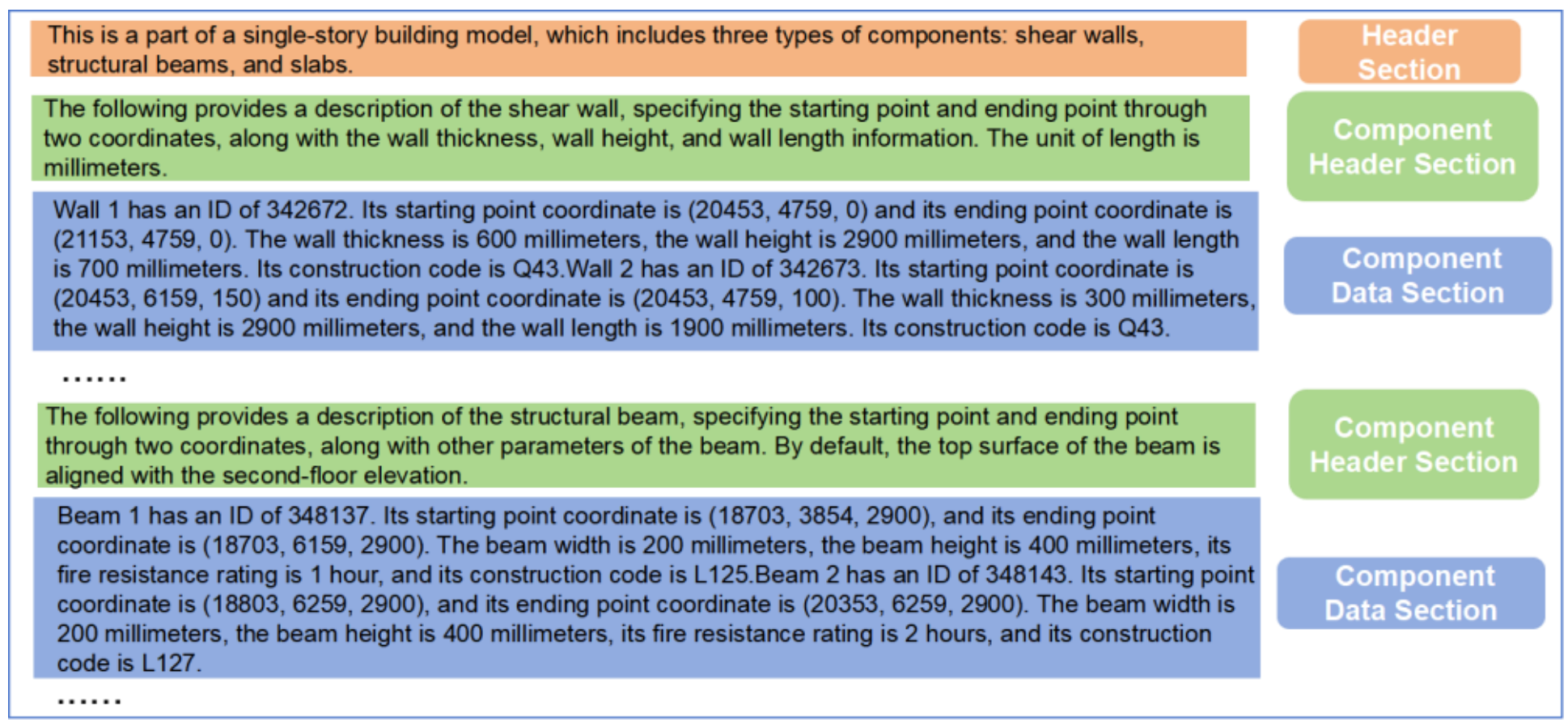

**Fig. 3.** BIM-to-Text Method and Example

Second, after textualization, the component descriptions are exported to a textual file for further usage. To ensure that the input length to the large model remains balanced and manageable, a fixed number of component data entries are randomly extracted from the file each time to form a data block. Although the components within a single block may not be spatially adjacent because they are evaluated independently, their number does not exceed a specified limit. This ensures a similar text length for different model parts, which could be further optimized by adjusting the number of components per block to boost the performance of LLMs.

### 3.2 Design defect identification

The BIM defect identification method based on the domain-specific LLM is illustrated in **Fig. 4**. First, the BIM model is converted into text and divided into chunks, each containing a balanced number of component descriptions. Then, defect identification rule prompts and output specification prompts are added to form the complete input for the LLM. Subsequently, the BIM domain-specific LLM is invoked to generate the defect identification results. Finally, the generated output is examined to determine whether hallucinations have occurred—if hallucinations are detected, the model regenerates the response; otherwise, the generated content is accepted as the final output. The accuracy of the defect identification results is verified through manual inspection.

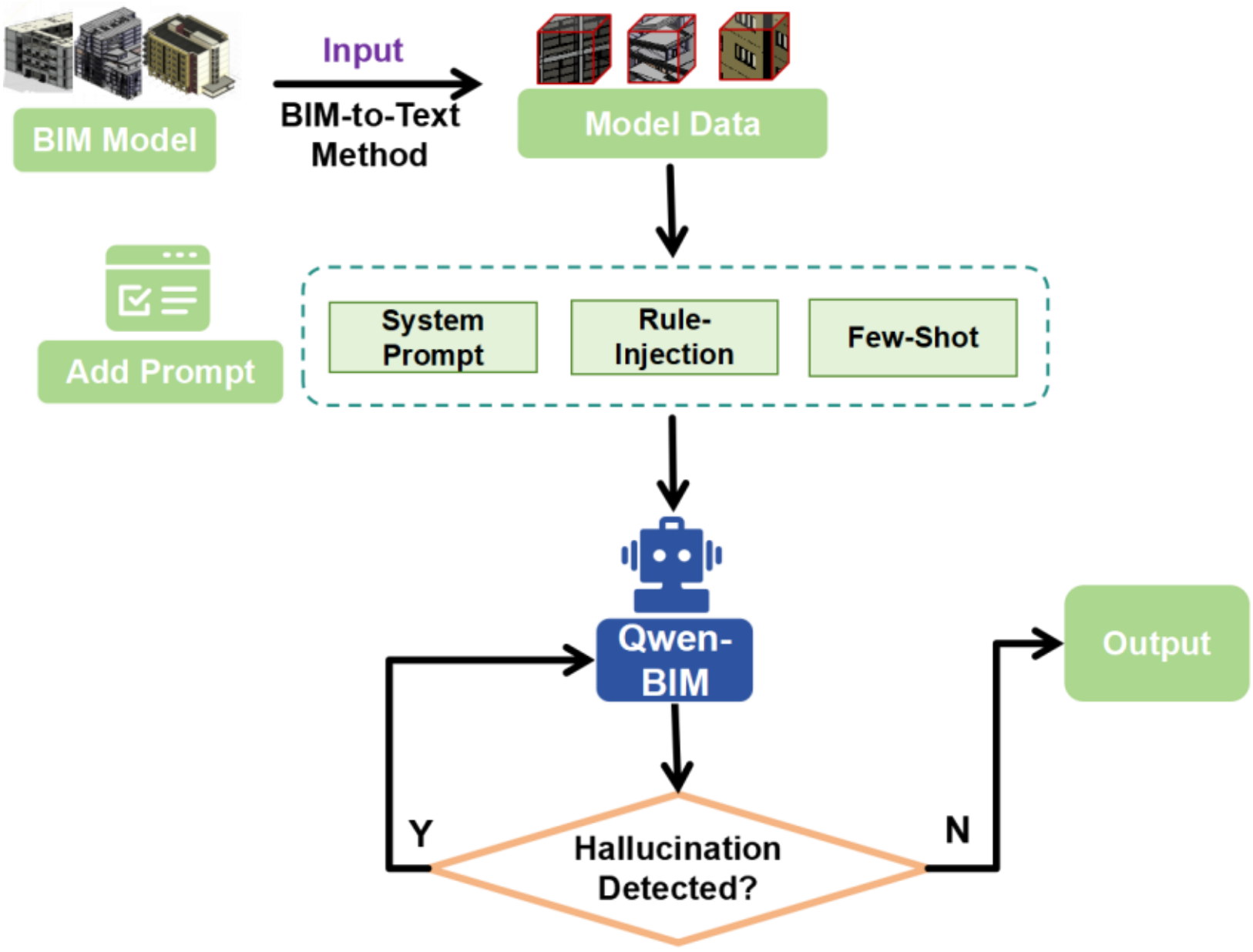

**Fig. 4.** BIM Defect Identification Method Based on Domain-Specific LLM and Prompt Learning

Based on the overall workflow, to ensure the accuracy and stability of the identification task, this study further designed a prompt framework for multi-category defect identification. Specifically, during the defect identification of each data chunk, the textual descriptions of individual components are concatenated into a single paragraph, to which system prompts, rule-injection prompts, and few-shot prompts are subsequently added.

(1) System Prompts

In prompt learning, system prompts define global rules or contextual settings, and role-playing helps guide the model to generate outputs aligned with specific professional contexts [59]. In this study, prompt design focuses on role specification, reasoning guidance, and output format constraints. A role-playing strategy is used to align responses with professional contexts, combined with chain-of-thought reasoning to enhance logical accuracy. The answer format and representative examples are explicitly defined to improve consistency and reduce randomness, while evaluation considers only the correctness of final answers.

After iterative design and experimental refinement, the unified system prompt used for model evaluation in this study is:

*"You are a professional architectural designer. In the following dialogue, after performing stepwise calculation and reasoning to obtain a result, please provide the final answer in the following format. Regardless of whether any reasoning steps are shown, you must add the string '[Final Answer]:' on a new line, and then, starting from the next line, give your answer to the question in complete sentences, without including any additional information or reasoning process. Please strictly follow the format shown in the example provided after the question when giving the final answer."*

(2) Rule-Injection Prompts

Task prompts are designed for the specific application scenario using the Few-Shot method

[60], incorporating natural language descriptions and explanatory examples of relevant code provisions, together with explicit inspection requirements. Through iterative experimentation, it was determined that expanding every provision individually is unnecessary; instead, combining structured representations with representative examples enables the model to understand rules more effectively, reduces input length, and lowers the likelihood of hallucinations. The rule-injection prompt adopted in this study is summarized in **Fig. 5**.

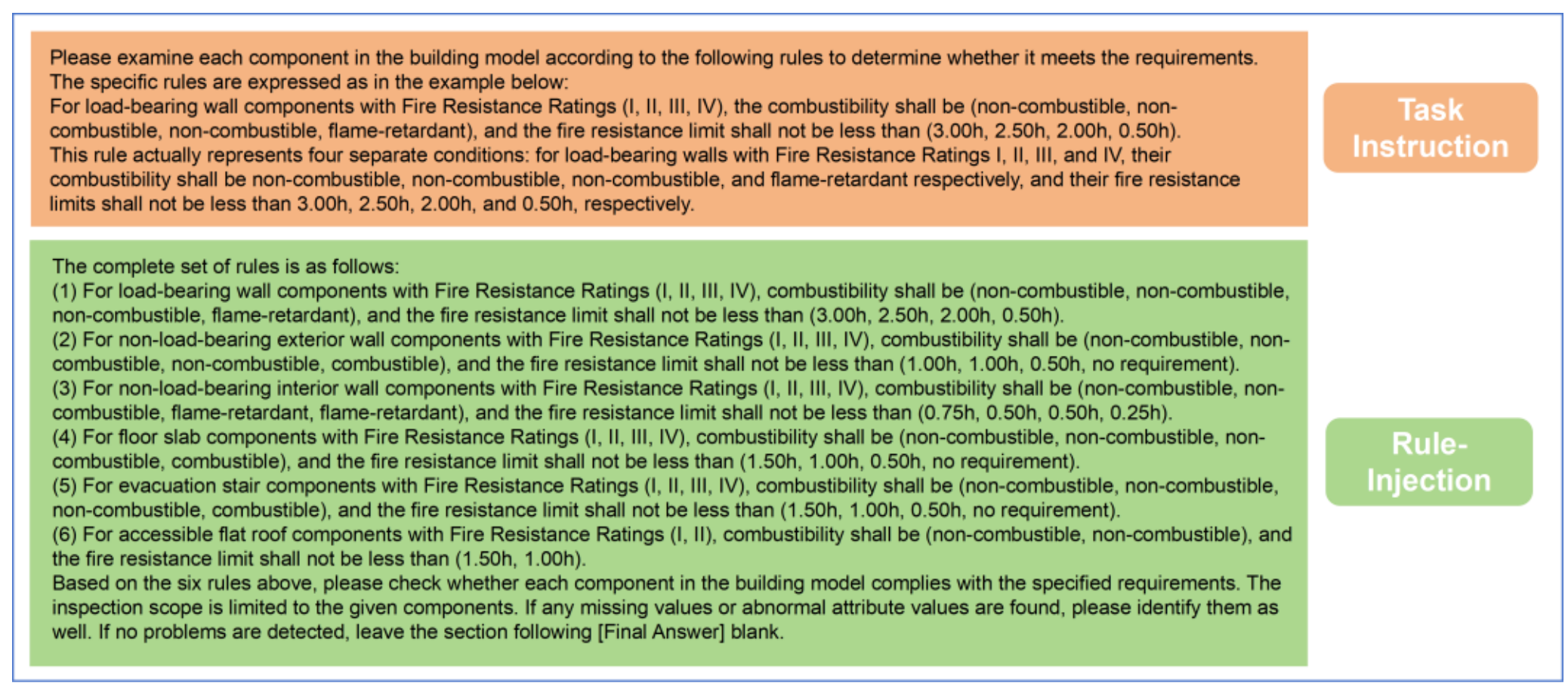

**Fig. 5.** Structure of the Rule-Injection Prompt

(3) Few-Shot Prompts

To further standardize the output format of the "Final Answer," this study employs few-shot example prompts that provide answer samples. By offering examples, the large model can generate responses that better align with user expectations. In this study, example prompts are appended after the test questions, following the format:

"Content following [Final Answer] should refer to the following format: + example text."

Here, "example text" represents the example corresponding to a specific test question.

The example generation process involves first selecting the answers for all test questions within a randomly chosen data block, then manually concealing part of the information and replacing it with "...", thus producing example texts for all test questions. For instance, for an area calculation problem of computational difficulty level 3, the example text is:

*"The ID of the first slab is 350353, and it is not rectangular. The ID of the second slab is 350380, and it is rectangular with an area of 1.47 m². ... Among them, the rectangular slab with the largest area has the ID ... and an area of ... m²."*

In this case, additional listed items, as well as the ID and area of the largest slab, are hidden.

Another type of hidden information concerns questions that require explanatory reasoning. For example, in Reasonableness Defect Identification Scenario 1, the example question is:

*"There are some suspicious wall thickness data in this part of the model. The wall thickness of the first wall is questionable; its ID is 342679, and its thickness is 20 mm. Because ..., this data may be incorrect."*

Here, the specific reason is concealed to prevent the example data from influencing the model's reasoning process and causing hallucination.

Moreover, specific examples are provided depending on the application scenario. For instance, in the identification of completeness and compliance defects concerning fire performance parameters, the following example prompt is used:

*"After [Final Answer], only list the ID of each problematic component and its specific issue. Separate the ID and the problem description with a comma, and separate different components with semicolons. For example: 348253, fire performance is combustible, not compliant; 350236, fire performance missing, not compliant."*

This prompt provides explicit examples and format requirements, thereby improving the quality and consistency of the model's output.

### 3.3 Defect Repair Suggestion Generation Based on RAG

To address the problem of BIM defect repair, this section proposes a defect repair suggestion generation method based on the domain-specific large model and RAG, as shown in **Fig. 6**. First, a construction method for the defect repair knowledge base is introduced to support the application of the RAG approach, ensuring the rationality of repair suggestions and improving generation quality. Then, based on the domain-specific large model and RAG, a fundamental method for generating BIM defect repair suggestions is established.

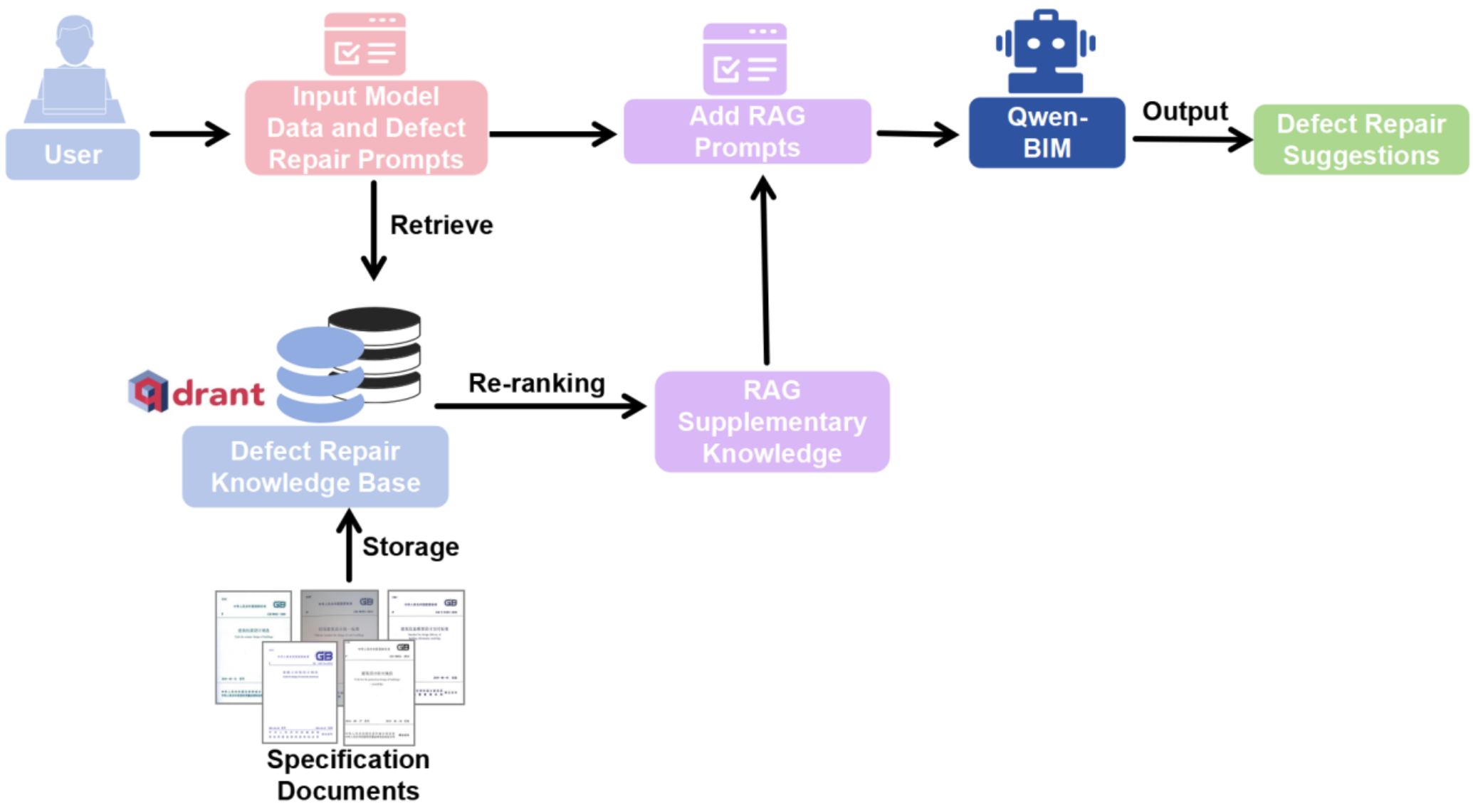

**Fig. 6.** Method for Generating Defect Repair Suggestions

#### 3.3.1 Knowledge Base for Defect Repair

To implement the RAG method, it is necessary to build a domain-specific knowledge base with semantic similarity retrieval capabilities, typically realized through a vector database. In this study, the open-source Qdrant vector database is employed to construct the defect repair knowledge base. Qdrant supports multiple similarity calculation methods and allows the attachment of textual payloads to vectors, thereby establishing a mapping relationship between vectors and corresponding text. In this study, each vector is assigned two payloads: a title and knowledge content.

As shown in **Fig.7**., the required repair knowledge differs depending on the type of defect: integrity and compliance defects require additional knowledge sources such as building codes and standards, whereas rationality defects are mainly associated with BIM design common sense, which has already been provided to the large model during fine-tuning and therefore does not require supplementary knowledge. Consequently, the defect repair knowledge base developed in this study primarily supports the repair of integrity and compliance defects, and its contents are mainly derived from code provisions.

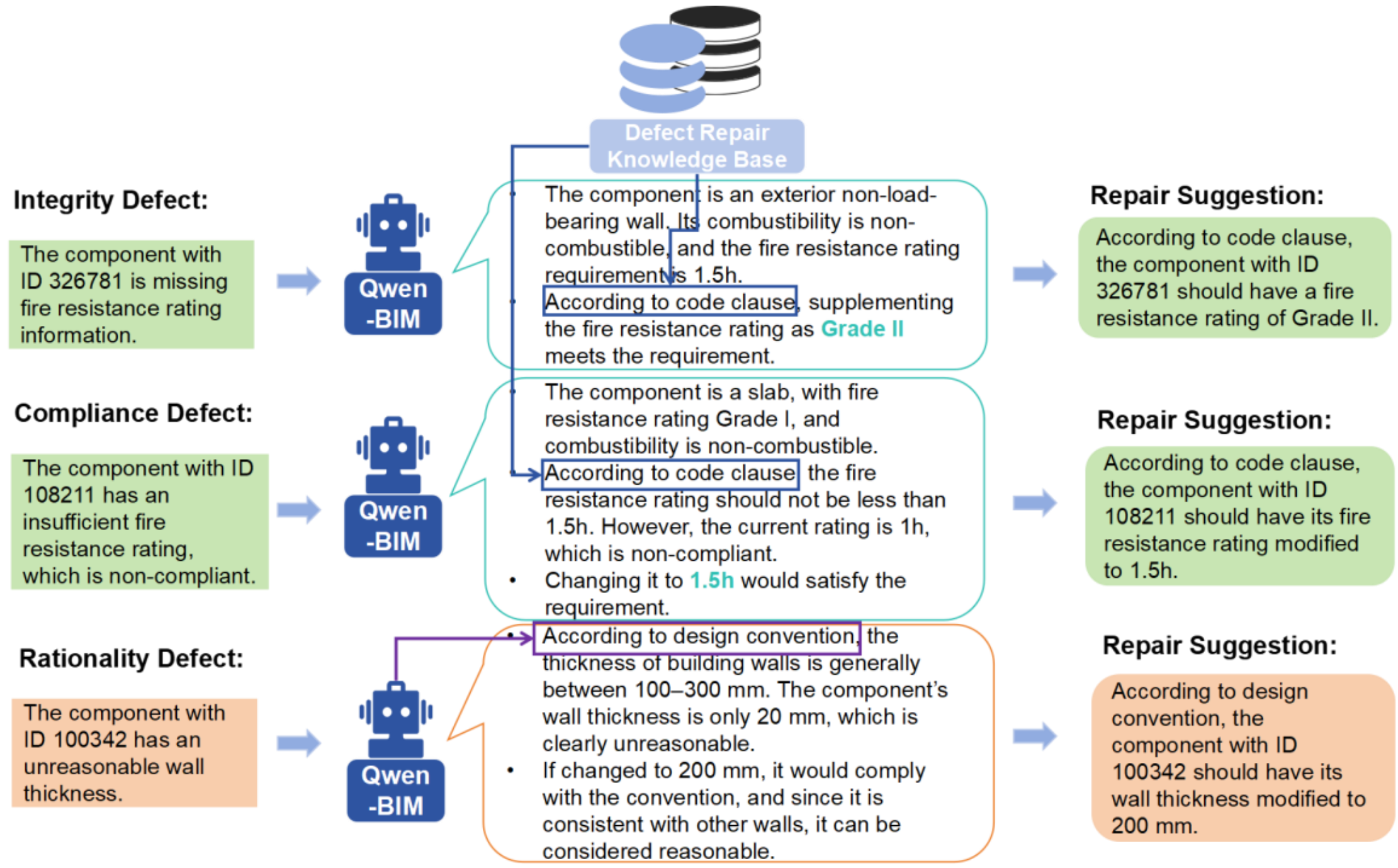

**Fig. 7.** Examples of Repair Suggestion Generation for Different Types of Defects

The construction process of the defect repair knowledge base is illustrated in **Fig. 8**. In this study, five key codes and standards were selected as domain knowledge sources, including "Unified Standard for Design of Civil Buildings" (GB 50352-2019), "Code for Design of Concrete Structures" (GB 50010-2010, 2015 edition), "Code for Fire Protection Design of Buildings" (GB 50016-2014, 2018 edition), "Code for Seismic Design of Buildings" (GB 50011-2010, 2016 edition), and "Standard for Design Delivery of Building Information Modeling" (GB/T 51301-2018).

Based on the content of these standards and relevant design documentation, a structured knowledge base was organized to support defect repair. The collected materials were organized into thematic chunks based on their content structure, with each chunk assigned a title combining relevant standard names and section information. Specifically, focusing on code provisions, an automated pipeline was utilized to extract texts, tables, and formulas into a hierarchical JavaScript Object Notation (JSON) format without arbitrary token truncation. These chunks were then encoded into 1024-dimensional vectors using the Embedding-3 model provided by Zhipu AI, and both the textual content and titles were stored as vectorized knowledge entries in the Qdrant database to support retrieval and reasoning in subsequent RAG processes. Considering that the semantic orientation of the normative texts in this study is relatively homogeneous, cosine similarity was deemed unsuitable for vector similarity calculation; therefore, Euclidean distance was adopted as the similarity metric.

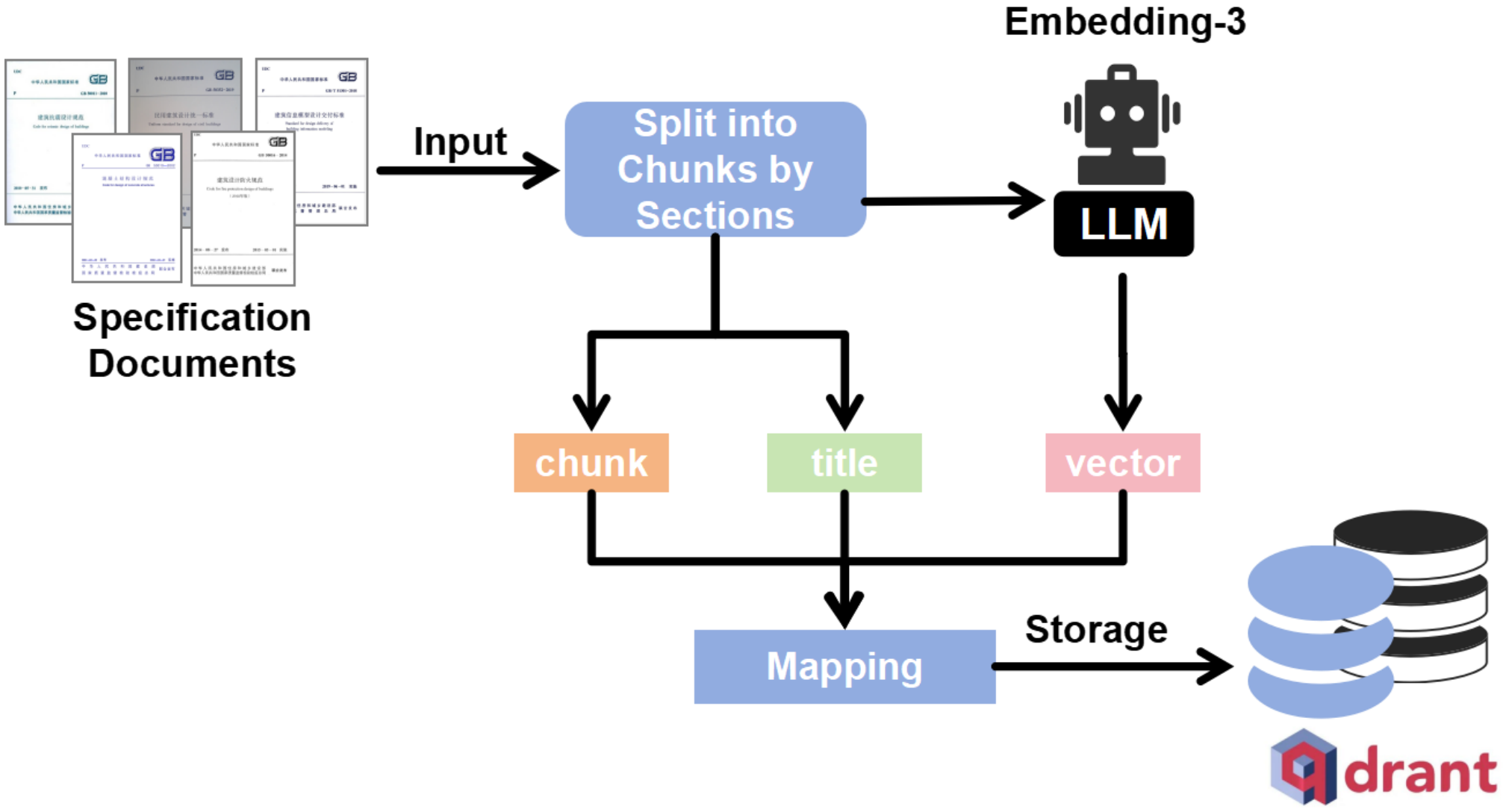

**Fig. 8.** Method for Constructing the Defect Repair Knowledge Base

The resulting defect repair knowledge base comprises 221 vectors, each mapped to a corresponding title and knowledge text. **Fig. 9** presents the principal component analysis (PCA) of the Qdrant database, where each point represents a vector. Data points from different standards exhibit clear clustering patterns. Beyond the specific application scenario in this study, the constructed defect repair knowledge base can also be applied to other defect repair tasks related to the aforementioned standards, demonstrating a certain degree of general applicability.

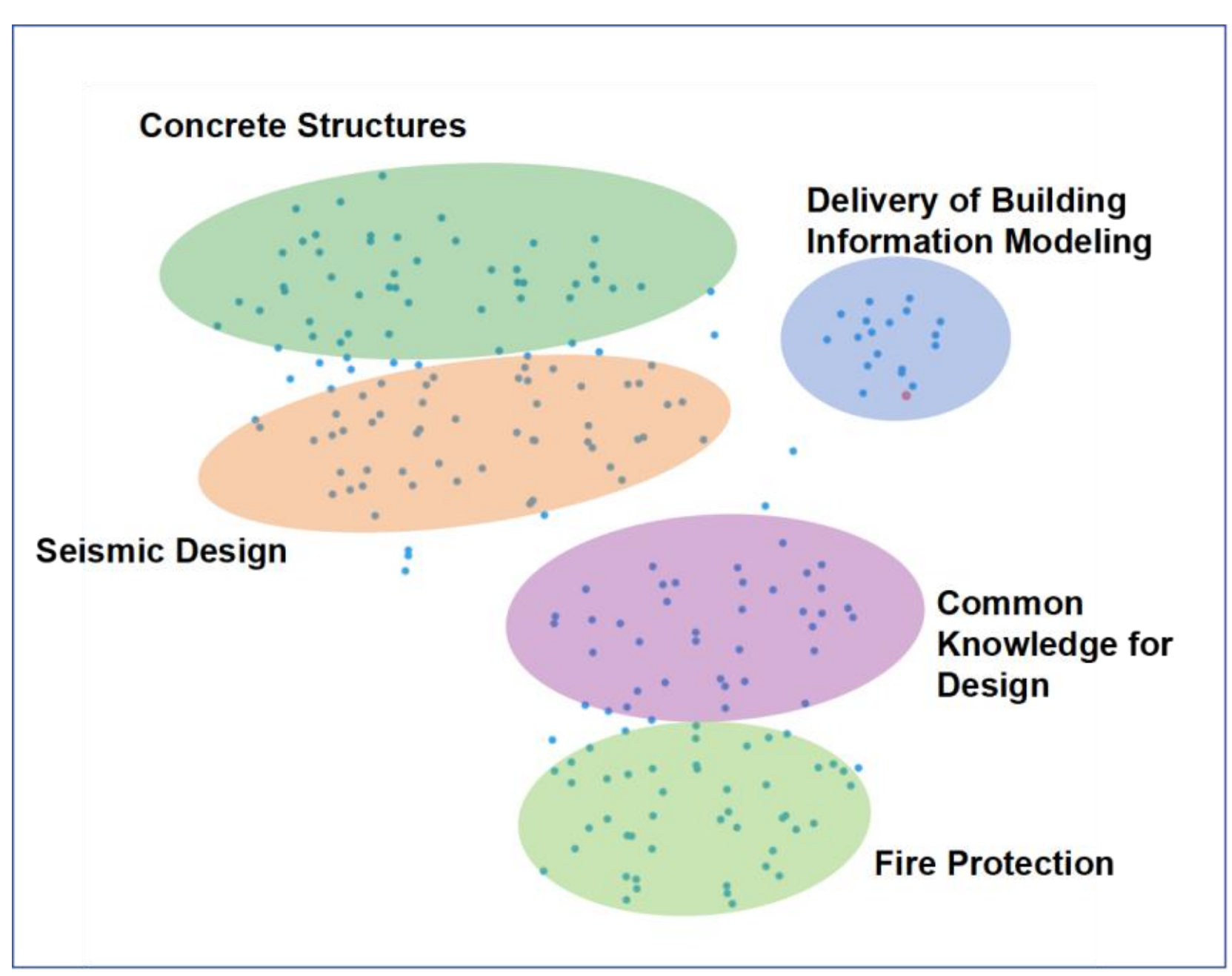

**Fig. 9.** Principal Component Analysis of the Defect Repair Knowledge Base

### 3.3.2 Method for Generating Defect Repair Suggestions

The method for generating defect repair suggestions based on RAG is illustrated in **Fig. 10**. First, all data chunks obtained through the chunking method described in Section 3.1 are traversed to locate those containing defective components. Then, based on the chunk data, component IDs, defect type, and detected defect information, an initial input for the large model is constructed. This input includes the system prompt, BIM-to-Text data, detected defect information, Rule-Injection Prompts, and Few-Shot Prompts. To incorporate RAG, the Qdrant database first retrieved the top ten text segments with the highest similarity scores from the defect repair knowledge base. These results were then re-ranked using the Rerank model provided by Zhipu ChatGLM to ensure semantic relevance, and the three most relevant segments were selected as the final RAG content. The retrieved regulatory texts were appended to the model input as external knowledge, and RAG-specific prompts were subsequently added. All inputs were then fed into the domain-specific large model.

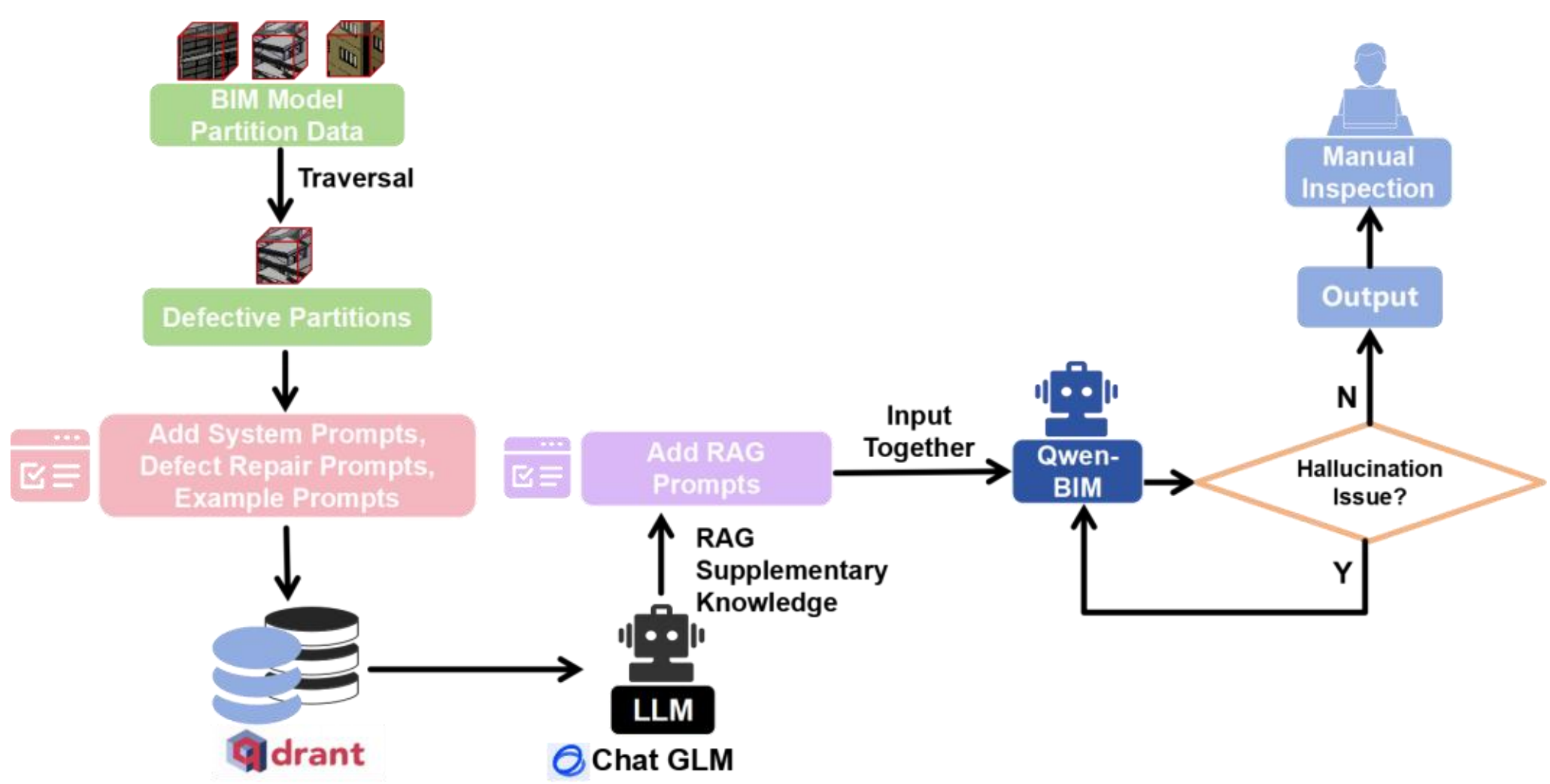

**Fig. 10.** Method for Generating Defect Repair Suggestions Based on RAG

The final complete input to the model consists of:

"Textualized model data + system prompt + defect repair prompt + few-shot prompts + RAG-retrieved knowledge + RAG prompt."

After the model generates the output, the presence of hallucinations is verified using the approach described in Section 3.4. If hallucinations are detected, regeneration is performed iteratively until a hallucination-free result is obtained. The rationality of the final repair suggestion is then confirmed through manual review.

The system prompt and few-shot prompts follow the same formulations as those in Section 3.2. The defect repair prompt must specify the location information of the component to be repaired (e.g., its ID), the specific defect (e.g., "insufficient fire resistance rating"), and the repair requirements, which may include referenced codes, standard names, or related details. Preliminary experimental tests show that these repair requirements do not need to be overly precise. For example, in fire protection code checking, the repair prompt does not need to include the full name of the code, since the defect description itself (such as "fire resistance limit") already contains sufficient key terms for the RAG system to retrieve the most relevant knowledge text.

The RAG prompt used in this study is as follows:

*"The following content consists of code provisions retrieved from the knowledge base based on the user's query. Please first determine whether these contents are relevant to the question, and then answer the user's question based on the relevant provisions."*

Each retrieved item is subsequently presented in segments, including its section title and corresponding text. Additionally, in practical applications, to ensure a moderate level of diversity in model outputs, the "temperature" parameter of the input is set to 0.5.

### 3.4 Hallucination Control Method

After obtaining the output from the large model, it is necessary to determine whether hallucination issues are present. A systematic analysis of the domain-specific large model Qwen-BIM revealed that hallucination still occurs during BIM-related tasks, mainly manifested in two forms: the model fabricates non-existent information beyond the input scope or falls into logical loops during reasoning. To address these types of hallucinations, this study designs the countermeasure strategy shown in **Fig. 11** and proposes two hallucination control methods.

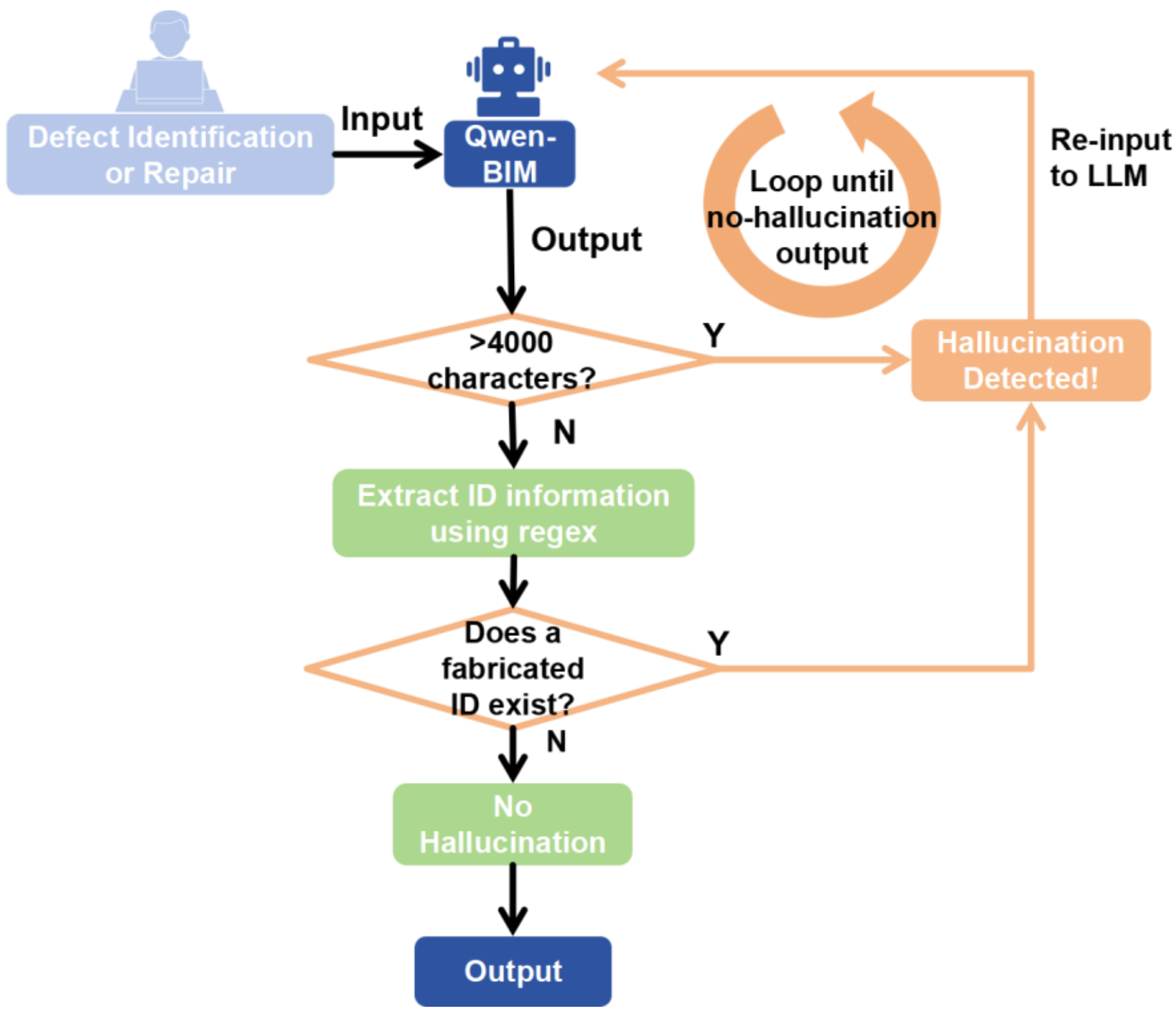

**Fig. 11.** Flowchart of the Hallucination Control Method

(1) As illustrated in **Fig. 12**, LLMs may produce hallucinated outputs that extend beyond the actual BIM data scope, such as fabricating nonexistent components (Wall 5–8) when only Wall 1–4 exist in the input. To mitigate such errors, a hallucination control method based on key identifier validation is proposed. In the final output, the study applies regular expressions to extract component IDs consisting of six or seven digits and compares them with the input data. If an ID not present in the input data appears, the model is deemed to have generated fabricated content. In addition to IDs, this study applies the same verification strategy to other key identifiers, including component names, attribute names, and type information, to ensure that these critical elements remain consistent between the input query and the model's output. This establishes a comprehensive identifier constraint that enhances the accuracy and coverage of hallucination detection.

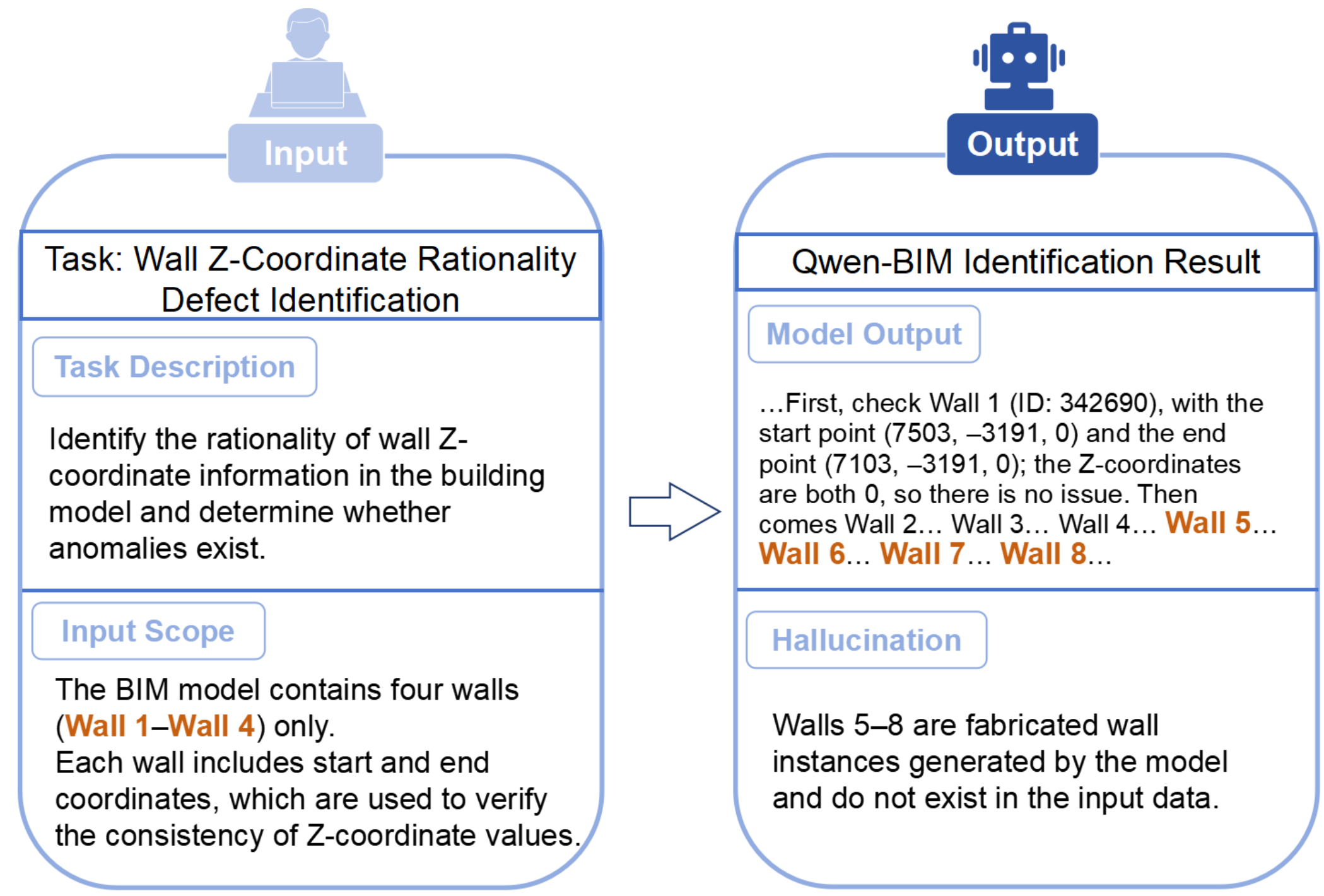

**Fig. 12**. Example of Fabricated Outputs in Qwen-BIM Defect Identification

(2) For infinite generation, a hallucination control method based on token-length threshold is adopted. When inputting data into the large model, the output upper limit is preset to 4,096 tokens. If hallucination occurs, the generated output typically approaches this length limit. Since the number of tokens roughly corresponds to the number of Chinese characters at a 1:1 ratio, the output text length is calculated, and when it exceeds 4,000 characters, hallucination generation is determined to have occurred.

Since hallucinations occur probabilistically and may persist across repeated inferences, this study introduces an iterative generation–validation mechanism applicable to both the defect identification and repair stages. The mechanism continuously evaluates model outputs against the hallucination control criteria; if any fabricated or inconsistent information is detected, the model is prompted to regenerate its response until a hallucination-free result is produced. This process ensures the reliability and robustness of outputs throughout the entire BIM defect identification and repair workflow.

### 3.5 Verification and Experiments

To systematically evaluate the proposed integrated framework, a comprehensive experimental design was established as illustrated in **Fig. 13**. The verification process begins with a residential building BIM model scenario where 100 design defects across integrity, rationality, and compliance categories were artificially injected based on fire protection codes. Subsequently, the validation is conducted across three distinct dimensions. First, the defect identification verification compares the accuracy of the traditional rule-based approach, the base LLM, and the domain-tuned Qwen-BIM. Second, the repair suggestion verification employs an ablation study to assess the generative performance of the models with and without RAG by evaluating the rationality rates of the output. Finally, the hallucination control verification tests the efficacy of the proposed reliability mechanism, which integrates key-identifier validation and token-length thresholds, in suppressing abnormal outputs and improving overall generation stability.

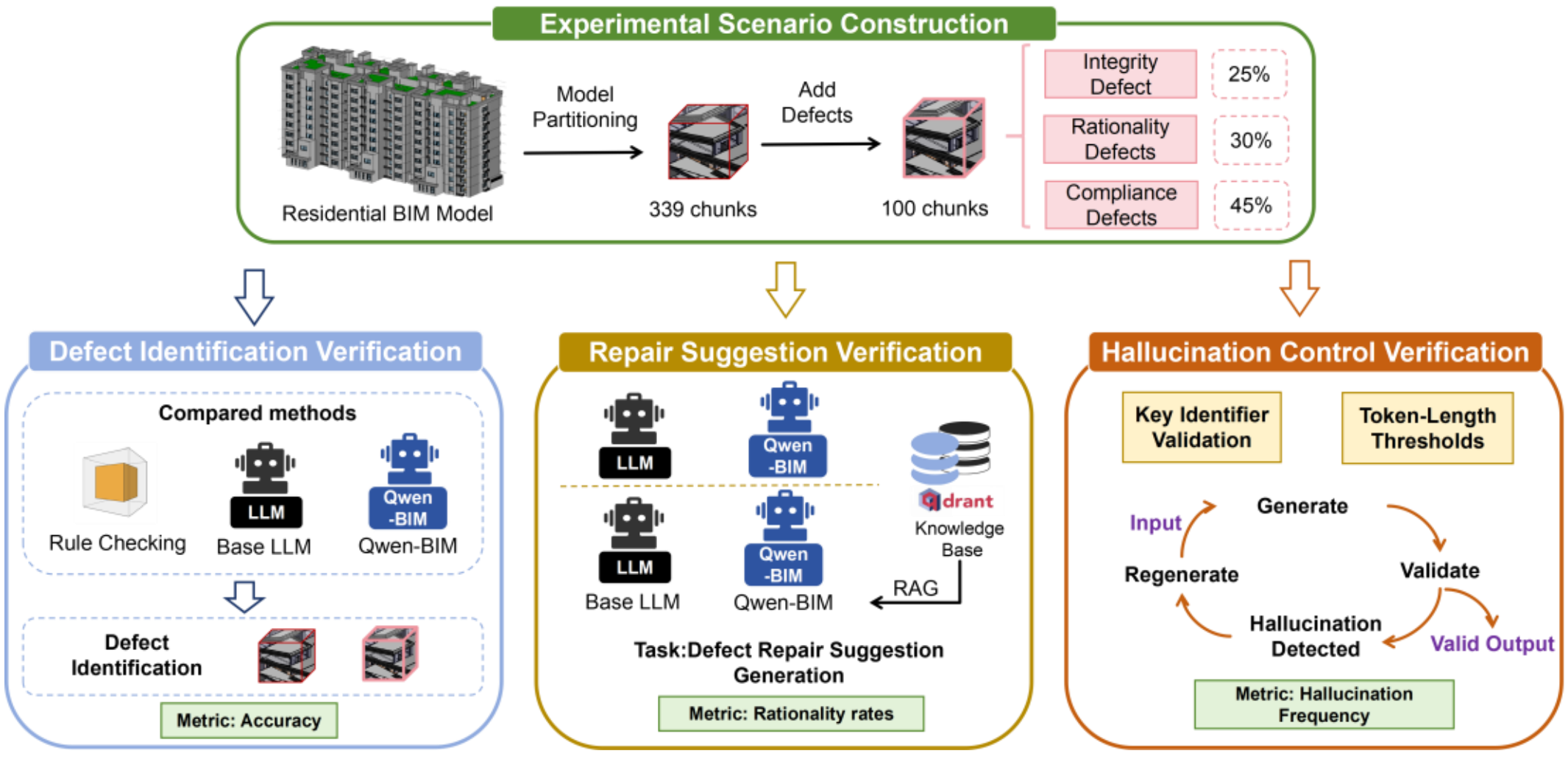

**Fig. 13**. Experimental design and verification of the integrated framework

#### 3.5.1 Experimental scenario construction

To verify the effectiveness of the proposed defect identification and repair methods, this study selected a specific application scenario for implementation and comparative analysis. In this scenario, the primary user requirement was to examine whether a residential building BIM model contained any violations of the “Code for Fire Protection Design of Buildings” (GB 50016-2014, 2018 edition) (hereinafter referred to as “the Code”). The BIM model, shown in **Fig. 14**, measures approximately 47 m in length (east–west), 18 m in width (north–south), and 32 m in height, excluding the underground structure. The BIM model primarily consists of six component types, including walls, slabs, evacuation stairs, roofs, doors, and windows, totaling 3,389 components to be checked. According to the chunking method described in Section 3.1, every ten components were grouped into one chunk, yielding 339 chunks in total. To strictly govern the generation process of the large language model, the 'temperature' parameter was set to 0.5 to maintain a moderate degree of randomness and prevent logical loops, while the maximum output token limit was constrained to 4096.

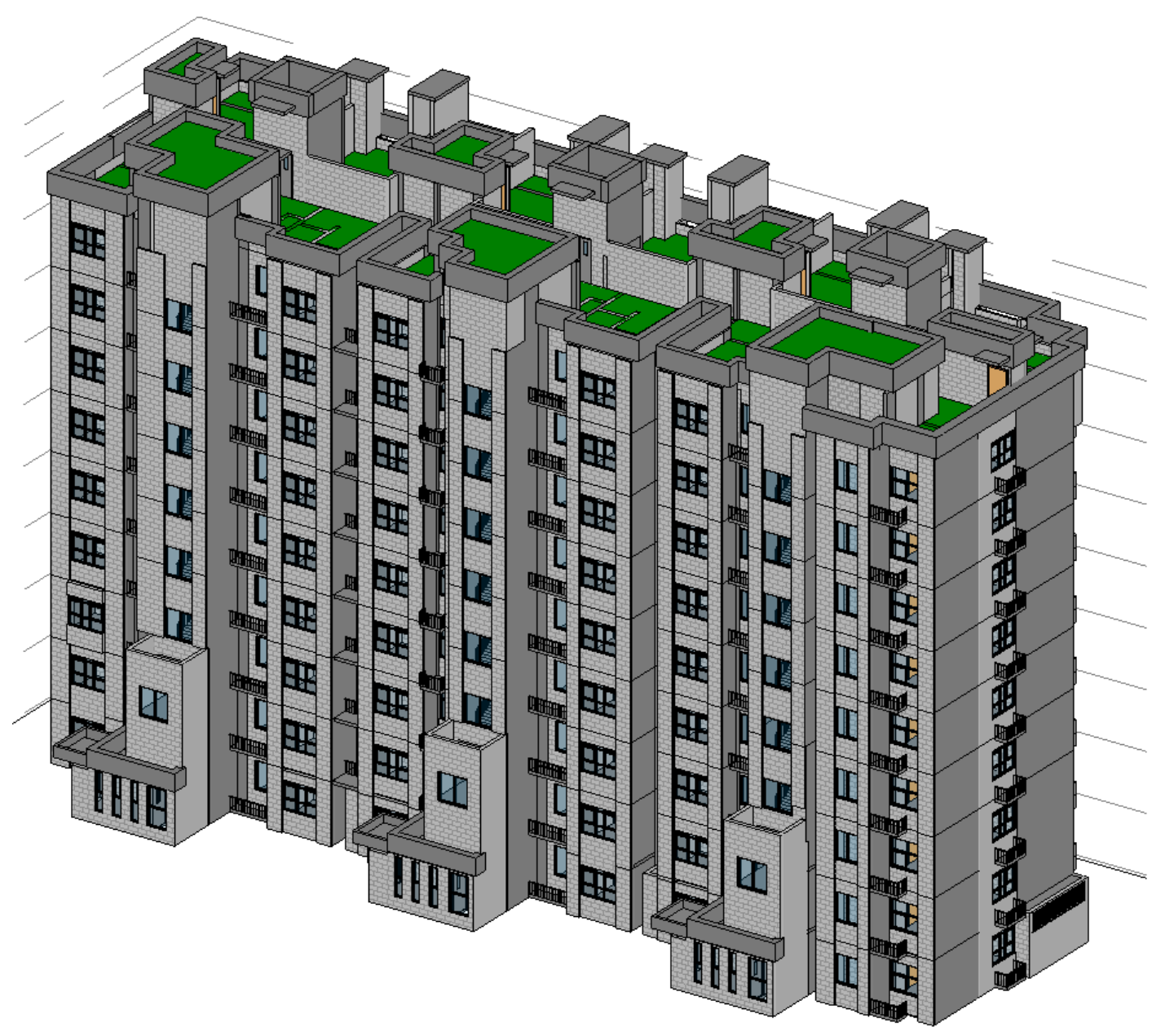

**Fig. 14.** Residential Building BIM Model Used for Experimental Verification

Based on the building type and major components, the inspection primarily focuses on the defects related the fire resistance rating, combustibility, and fire resistance limit for relevant elements. Since the Code presents these requirements in tabular form, they were converted in this study into a rule-based textual representation to facilitate automated checking. The rule expressions were further refined based on the component categories and attributes present in the BIM model. **Table 2** presents the converted results. Because the Code does not specify corresponding requirements for doors and windows, these components were excluded from this experiment.

**Table 2** Converted Code Provisions

| Component Type | Clause Content |
| --- | --- |
| Wall | For buildings of Fire Resistance Rating I/ II/ III, load-bearing wall components shall be non-combustible, and their fire resistance limit shall not be less than 3.00/ 2.50/ 2.00 hours. |
| | For buildings of Fire Resistance Rating IV, load-bearing wall components shall be flame-retardant, and their fire resistance limit shall not be less than 0.50 hours. |
| | For buildings of Fire Resistance Rating I/ II/ III, non-load-bearing exterior wall components shall be non-combustible, and their fire resistance limit shall not be less than 1.00/ 1.00/ 0.50 / hour. |
| | For buildings of Fire Resistance Rating IV, non-load-bearing exterior wall components shall be combustible. |
| | For buildings of Fire Resistance Rating I/ II, non-load-bearing interior wall components shall be non-combustible, and their fire resistance limit shall not be less than 0.75 /0.50 hours. |
| | For buildings of Fire Resistance Rating III/ IV, non-load-bearing interior wall components shall be flame-retardant, and their fire resistance limit shall not be less than 0.50/ 0.25 hours. |
| Floor Slab | For buildings of Fire Resistance Rating I/ II/ III, floor slab components shall be non-combustible, and their fire resistance limit shall not be less than 1.50 / 1.00/ 0.50 hours. |
| | For buildings of Fire Resistance Rating IV, floor slab components shall be combustible. |
| Stair | For buildings of Fire Resistance Rating I/ II/ III, evacuation stair components shall be non-combustible, and their fire resistance limit shall not be less than 1.50 hours. |
| | For buildings of Fire Resistance Rating IV, evacuation stair components shall be combustible. |
| Roof | For buildings of Fire Resistance Rating I, accessible flat roofs shall be non-combustible, and their fire resistance limit shall not be less than 1.50 hours. |
| | For buildings of Fire Resistance Rating II, accessible flat roofs shall be non-combustible, and their fire resistance limit shall not be less than 1.00 hour. |

In the original BIM model, relevant defects were relatively few. Therefore, this study first generates 100 defects and recorded the component IDs and defect categories via Revit Application Programming Interface (API). The defects were classified into three types: integrity defects (25 instances), rationality defects (30 instances), and compliance defects (45 instances).

Integrity defects refer to missing component information, such as fire resistance rating, combustibility, or fire resistance limit. Rationality defects simulate common modeling errors where attribute data are complete but contain unreasonable values—for instance, assigning a non-existent "Level V" fire rating, using Arabic numerals instead of Chinese numerals, or entering excessive fire resistance limits (e.g., 300 hours). Compliance defects, in contrast, indicate that component attributes fail to meet relevant code requirements, such as non-compliant combustibility or insufficient fire resistance limits.

### 3.5.2 Comparative experiments on defect identification

To verify the effectiveness of the proposed defect identification framework, comparative experiments were conducted using the defect scenarios constructed in Section 3.5.1, involving three methods: the traditional rule-based checking approach, the base LLM Qwen2.5-14B-Instruct (before fine-tuning), and the domain-tuned Qwen-BIM model.

The rule-based approach was implemented using the Revit Model Checker, an official Autodesk tool for automated rule verification. In this process, relevant code provisions were encoded into Extensible Markup Language (XML)-based rule files, where each rule defined its inspection objects, constraint conditions, and filtering criteria [61]. These rule files were configured and executed within the Model Checker environment to perform automatic defect inspection [62].

For the proposed approach, both Base LLM and Qwen-BIM adopted the defect identification method developed in this study, which integrates prompt learning with a hallucination control mechanism to improve identification accuracy and reliability. The same BIM model and defect scenarios were used for all three methods to ensure experimental consistency. Identification accuracy was used as the primary evaluation metric to quantitatively compare the performance of different approaches.

#### 3.5.3 Verification of the Defect Repair Suggestion Generation

In this section, the method proposed in Section 3.3 was applied to automatically generate repair suggestions for 100 defects, including integrity, compliance, and rationality defects, in order to evaluate the generation performance of the domain-specific large model Qwen-BIM and the base model Qwen2.5-14b-instruct under both RAG and non-RAG conditions, with accuracy adopted as the evaluation metric for comparison.

The rationality of the generated repair suggestions was evaluated based on the following criteria: for combustibility modifications or additions, the generated text had to be fully consistent with the code descriptions to be considered reasonable; for fire resistance rating and fire resistance limit modifications, a deviation within two levels or less than two hours was considered acceptable. For instance, if the required fire resistance limit was 1 hour, a model output of 1 hour would be regarded as reasonable, whereas a suggestion of 4 hours would be considered unreasonable due to the excessive deviation.

#### 3.5.4 Verification of the Hallucination Control

Based on the hallucination control strategy proposed in Section 3.4, which combines key identifier validation and token-length threshold, this experiment was designed to evaluate its effectiveness in suppressing abnormal outputs and enhancing generation stability in defect identification tasks. The base LLM Qwen2.5-14B-Instruct and the domain-tuned Qwen-BIM were selected as test models, and the BIM model scenario constructed in Section 3.5.1 was used as the experimental context.

During the experiment, both models performed multiple rounds of defect identification using identical inputs. The output of each round was examined through the hallucination control mechanism to detect any hallucinated content; if hallucination was detected, the model was prompted to regenerate until a hallucination-free result was achieved. Hallucination frequency and identification accuracy were adopted as the primary evaluation metrics to quantify the effectiveness of the proposed control method.

## 4 Results and Discussion

### 4.1 Defect Identification

#### 4.1.1 Performance of Defect Identification

Based on the experimental design described above, this section further compares the performance of the traditional rule-based approach with the proposed method that combines domain-specific LLMs and prompt learning, and analyzes the difference in defect identification accuracy between the domain model Qwen-BIM and the base model Qwen2.5-14b-instruct.

**Fig. 15** shows the accuracy of defect identification under the three methods for the 100 artificially introduced defects , where accuracy refers to the proportion of components for which both the ID and defect type were correctly identified. The rule-based method can efficiently check all components in a BIM model and output defect lists. However, since it relies on strictly logical rule statements, it is inherently incapable of processing rationality defects, often misclassifying them as compliance issues or ignoring them entirely. In contrast, the large-model-based approach leverages semantic interpretation to better identify such context-dependent issues, although its inherent stochasticity makes it prone to generating hallucinations, which may slightly reduce its accuracy in identifying deterministic integrity and compliance defects.

Driven by these underlying mechanisms, the experimental results indicate that the rule-based method achieved an overall accuracy of 70%, while the domain-specific model Qwen-BIM achieved a higher accuracy of 85%. To evaluate this performance objectively, the results should be interpreted separately based on the defect categories. On the compliance and integrity defects that the rule-based system can handle, the LLM-based approach slightly outperforms traditional rule checking. The 15% overall accuracy improvement is primarily attributed to the exclusive ability of the model to address rationality defects, accounting for 30% of the test set. Therefore, this accuracy gap objectively reflects a significant broadening of the inspection scope rather than a mere accuracy increment on identical tasks.

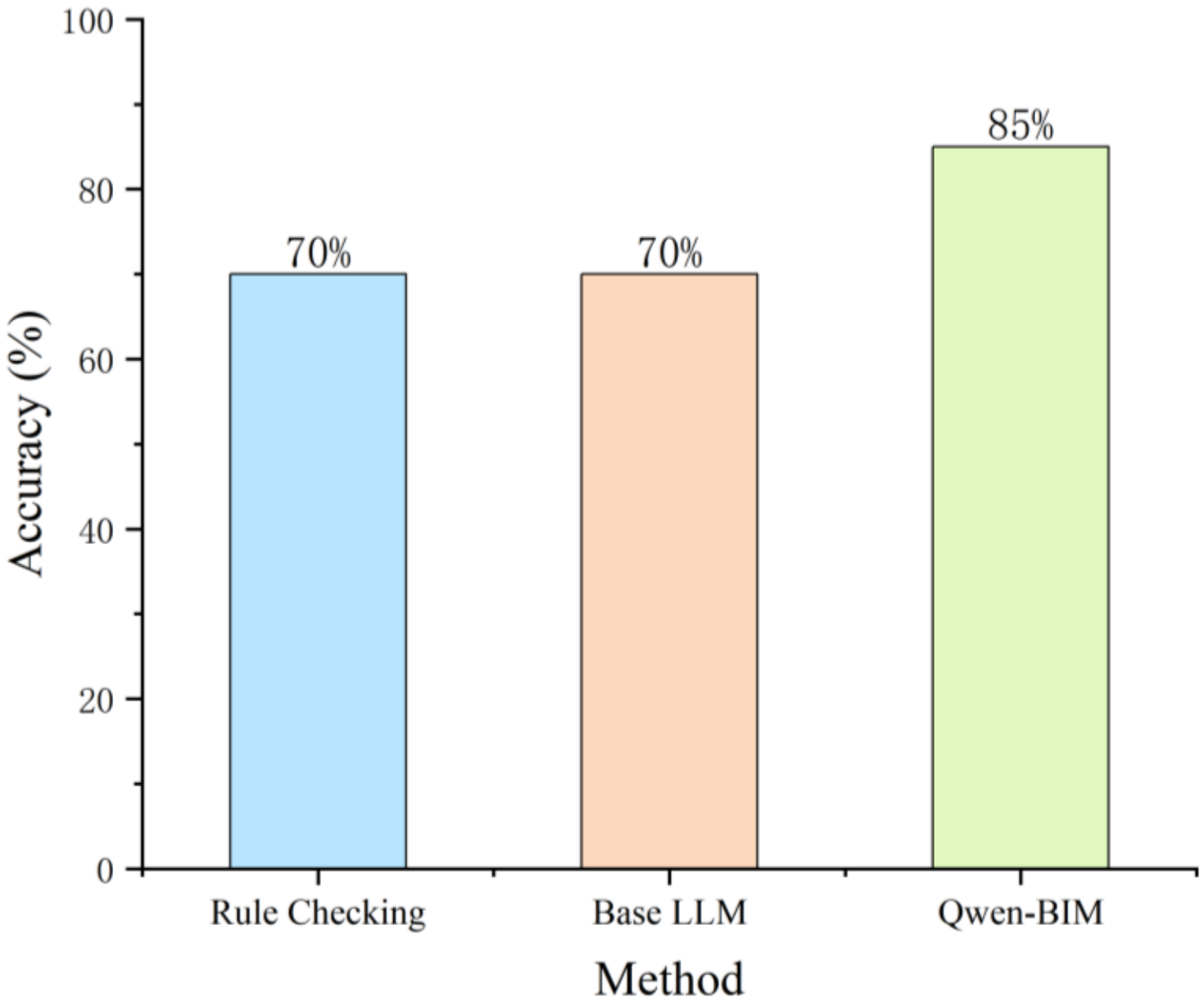

**Fig. 15.** Accuracy of defect identification using different methods

**Fig. 16** presents the identification accuracy of the base model and Qwen-BIM for different defect categories. The large-model-based method performs well in detecting rationality defects and can tolerate occasional noisy or incomplete information. Moreover, the figure shows that the fine-tuned domain model Qwen-BIM outperformed the base model in identifying compliance and integrity defects, though its accuracy for rationality defects slightly decreased. This is because the fine-tuning process emphasized reasoning ability, requiring higher-quality reasoning-related data. The fine-tuning dataset, constructed from normative texts and Revit-related web resources, may contain uneven data quality, leading to relatively weaker learning of BIM-specific design knowledge. Future research could focus on improving the quality of these fine-tuning datasets to further enhance the accuracy of multi-type defect identification.

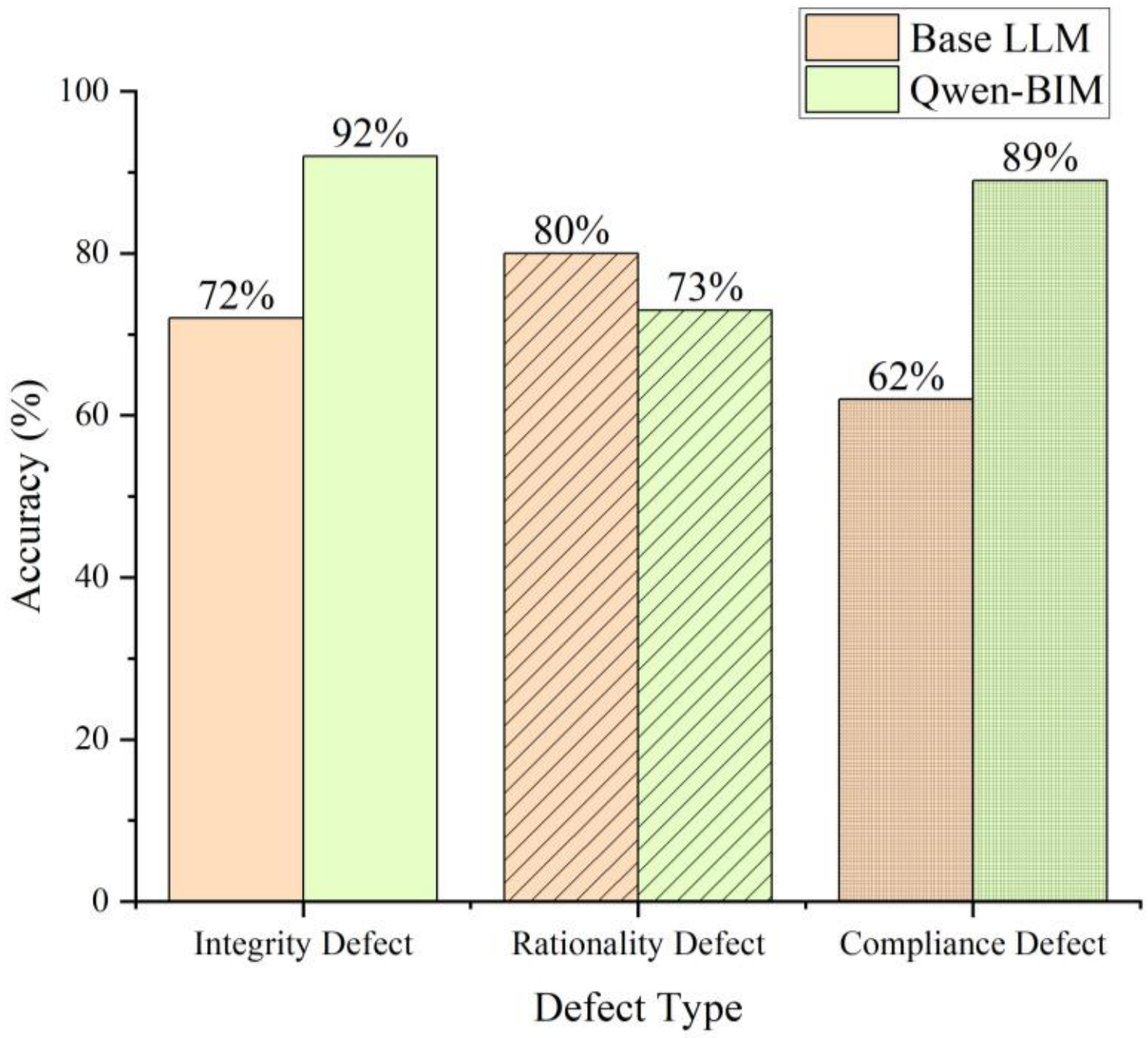

**Fig. 16.** Identification accuracy for different defect categories

#### 4.1.2 Discussion of Typical Cases

To verify the effectiveness of the Qwen-BIM-based inspection method combined with prompt learning, four representative cases are analyzed, as shown in **Fig. 17**.

In Case 1, the wall component is labeled as "Class A non-combustibility," which semantically matches "non-combustible." The Model Checker, relying on exact string matching, reports non-compliance, while Qwen-BIM correctly identifies the equivalence and judges the component as compliant.

In Case 2, the stair component is described as "non-combustible," whereas the code specifies "non-combustibility." The Model Checker again misclassifies the case, whereas Qwen-BIM recognizes the adjective–noun correspondence and provides the correct result.

In Case 3, the wall component's fire-resistance rating is recorded as "1" instead of "Grade 1." The Model Checker fails to perform the check due to rule mismatch, while Qwen-BIM interprets

“1” as “Grade 1” and completes the inspection correctly.

In Case 4, the slab component shows a fire-resistance limit of 30 hours, which satisfies the code numerically but is unrealistic in practice. The Model Checker overlooks this anomaly, while Qwen-BIM identifies the over-specification and issues a rationality warning, consistent with contextual patterns in the training data.

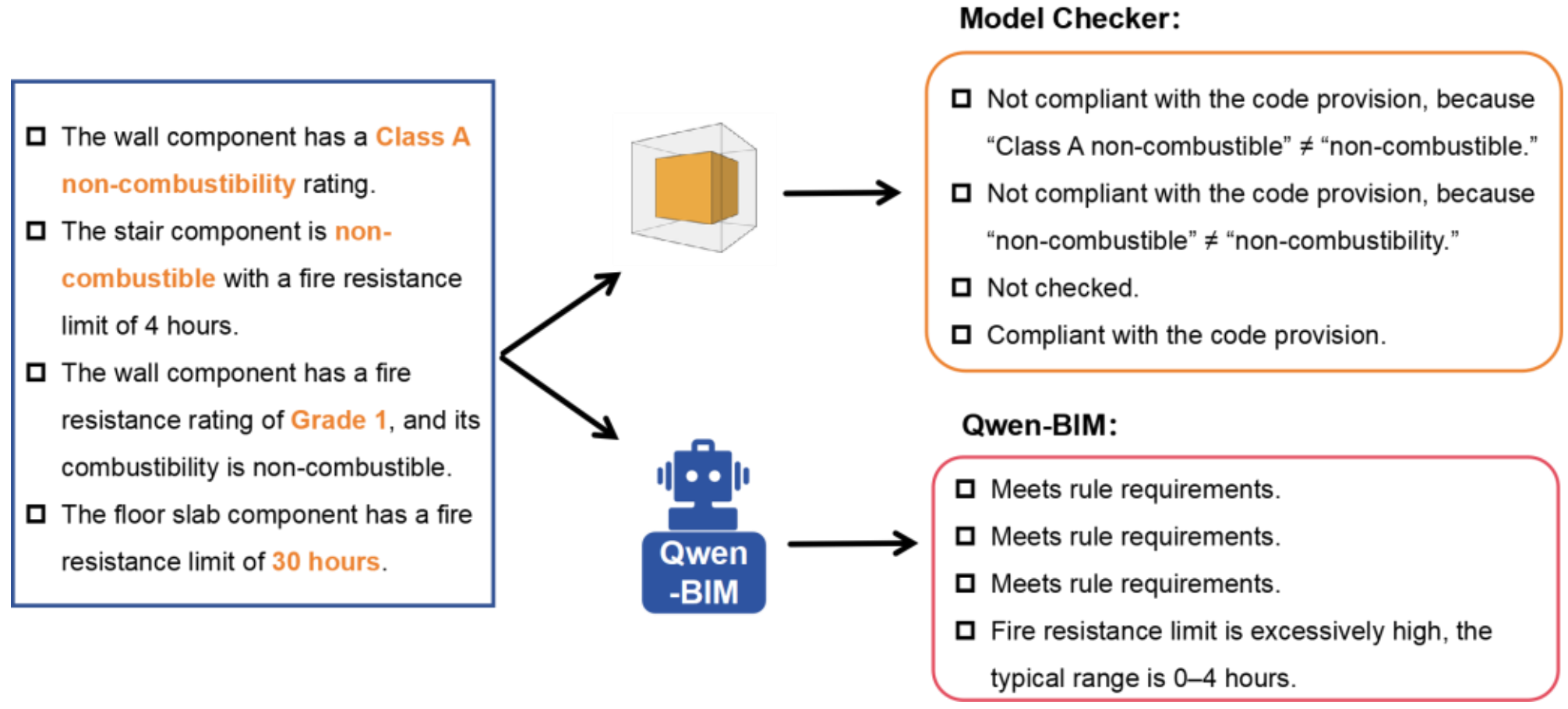

**Fig. 17.** Typical cases comparing Qwen-BIM and Model Checker in semantic and rationality inspection

In summary, the method based on the domain specific large language model effectively expands the capability to identify rationality issues. The model leverages its training data to recognize the semantic consistency of different expressions and detect gross deviations from normative value ranges. Furthermore, the proposed method demonstrates the capability to simultaneously process multiple types of problems including integrity, rationality, and compliance defects within the tested scenario. This highlights a key value in capability expansion where the approach slightly outperforms traditional rule checking on compliance and integrity defects that rules can handle, while adding a new capability for rationality checking that purely syntactic rules lack, thereby effectively complementing traditional methods.

However, the experiments also revealed two current limitations of the approach: (1) relatively low generation efficiency, and (2) the presence of hallucination phenomena. Future research can explore quantization and other acceleration techniques to improve model response speed, as well as employ larger-parameter models to enhance long-text understanding and reduce hallucination occurrences.

### 4.2 Defect Repair Suggestion Generation

**Fig. 18** illustrates the rationality rates of repair suggestions generated for 100 defects by both the base model and Qwen-BIM, with and without the integration of RAG. The results show that the introduction of RAG significantly improved the rationality of repair suggestions for both models, with Qwen-BIM’s rationality rate increasing from 79% to 94%. This demonstrates that the retrieval-augmented mechanism centered on normative provisions plays a critical role in enhancing the rationality of repair suggestions, achieving improvements beyond those gained through fine-tuning alone. In addition, Qwen-BIM consistently outperformed the base model overall, confirming the positive impact of fine-tuning on the model's understanding of domain-specific semantics.

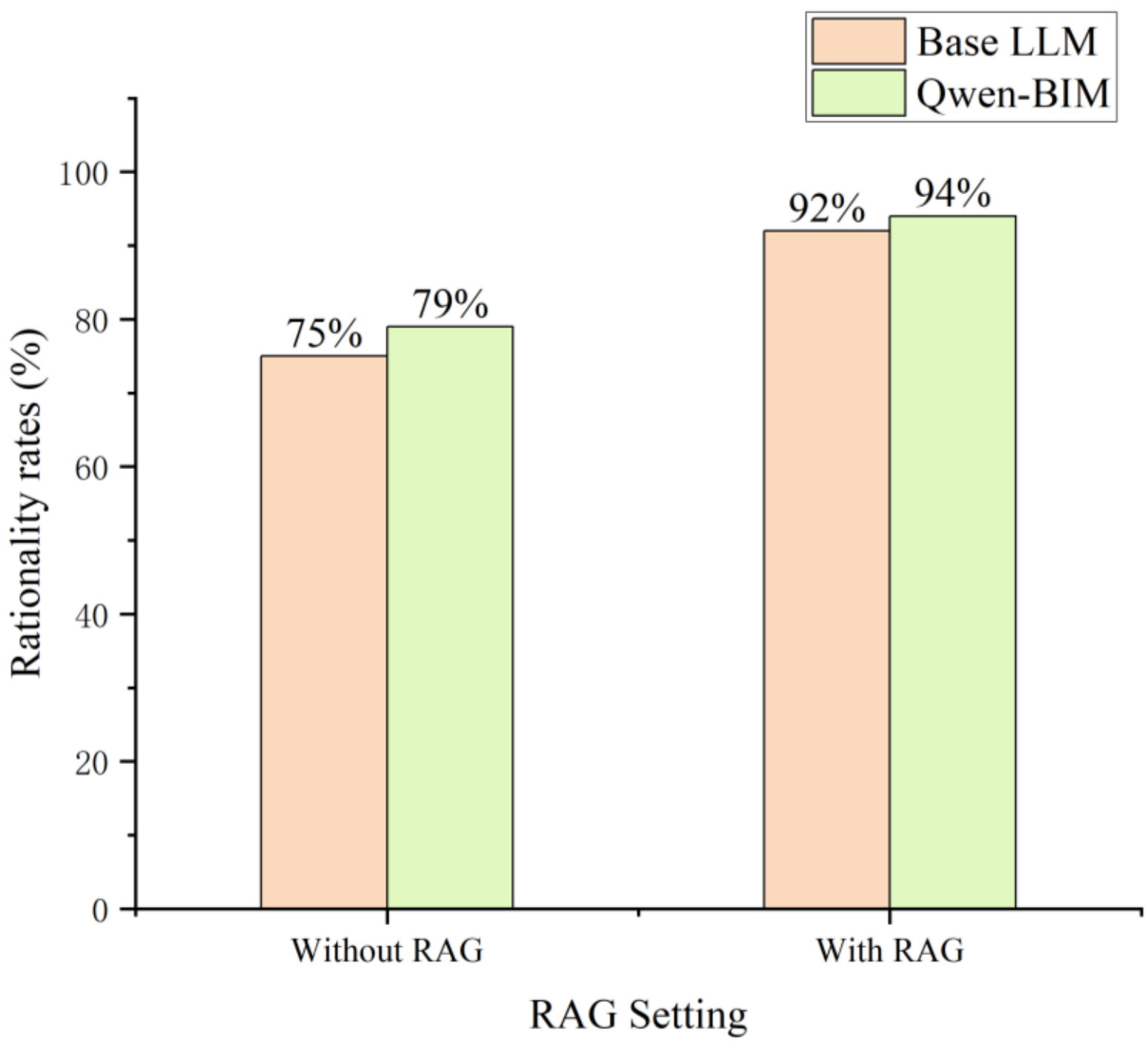

**Fig. 18.** Rationality rates of defect repair suggestions based on RAG

In addition, hallucination phenomena occurred only in Qwen-BIM when RAG was used, with a total of four cases observed. Further analysis of the model's reasoning process, as shown in **Fig. 19**, revealed that when no RAG knowledge was provided, Qwen-BIM tended to rely on commonsense reasoning or infer missing or non-compliant information by referencing other component data. This approach involved shorter reasoning chains but, due to the uncertainty of commonsense knowledge, often produced less accurate or unreasonable suggestions. Conversely, when RAG information was introduced, the model performed step-by-step analysis based on normative texts. Although this could occasionally introduce minor hallucination issues, it substantially improved the rationality and reliability of the generated repair suggestions.

**Defect Repair Problem of the Same Sample**

**Without RAG:**
**Tends to rely on common sense and analyze other components.**

Okay, I now need to address the issue raised by the user regarding component defects in the building model. First, I need to carefully review all the information provided by the user to ensure that the details of each component are correctly understood...
...For the fourth and fifth components, with IDs 350838 and 350925, the fire resistance limits are insufficient. According to the examples provided by the user, it may be necessary to adjust the fire resistance limits. For instance, **other walls have fire resistance limits of 4 hours or 3.75 hours**, while these two walls have a fire resistance limit of only 0.3 hours, which is clearly non-compliant. It may be necessary to change the fire resistance limit to 4 hours in order to meet the requirements of Grade II fire resistance. However, it must be confirmed whether all exterior walls are required to have a fire resistance limit of 4 hours. **According to the code, building components with a Grade II fire resistance rating generally require a fire resistance limit of 4 hours**, therefore these two walls should have their fire resistance limits changed to 4 hours...
[Final Answer]: ... For component 350838, the fire resistance limit should be changed to 4 hours...

**Using common sense leads to errors**

**With RAG:**
**Tends to rely on codes.**

Okay, I now need to handle the defects of each component in the building model provided by the user. First, I need to carefully review all the information provided, including each component's ID, the specific issues, and the requirements in the code knowledge base. Then, I need to analyze each issue one by one and propose repair solutions. **I must determine whether the combustibility and fire resistance limits of each component comply with the requirements of the code knowledge base. First, I need to identify the type of each component (interior wall, exterior wall) and whether it is load-bearing, and then determine the combustibility and fire resistance requirements according to its fire resistance rating...**
...The fourth and fifth components, with IDs 350838 and 350925, have insufficient fire resistance limits. Both are exterior walls made of aerated concrete blocks, non-load-bearing, with a Grade II fire resistance rating. **According to the code, non-load-bearing exterior walls with a Grade II fire resistance rating should have a fire resistance limit of at least 1.00 hour**, while both of these components have only 0.3 hours, which is clearly insufficient. Therefore, their fire resistance limits need to be adjusted to a compliant value, for example, 1.00 hour.
[Final Answer]: ... For component 350838, the fire resistance limit should be changed to 1.00 hour...

**strictly relying on codes yields correct answers**

**Fig. 19.** Comparison of Qwen-BIM's performance with and without RAG integration

### 4.3 Evaluation of Hallucination Control

A statistical analysis was conducted on the hallucination phenomena observed during defect identification using both the base LLM and Qwen-BIM. As shown in **Fig. 19**, the figure presents the frequency of hallucination occurrences after applying the proposed hallucination control method. The term "First" refers to the total number of chunks (out of 339) that exhibited hallucinations during the initial round of defect identification, while "Second" denotes the number of those same chunks that continued to produce hallucinations upon re-identification, and so forth. The "Total" represents the cumulative number of hallucination occurrences across all identification rounds.

The results demonstrate that the proposed hallucination control strategy effectively constrains abnormal outputs generated by LLMs. For Qwen-BIM, a total of 40 hallucinated chunks were detected in the first round; after re-identification with the hallucination control mechanism, only 3 chunks still exhibited hallucinations, indicating that 92.5% of hallucinations were eliminated after a single control iteration. Following four control cycles, hallucinations were completely removed, confirming the method's effectiveness in suppressing cumulative hallucination and improving generation stability within limited iterations. For the base model, a similar decreasing trend was observed under the same control mechanism, suggesting that the proposed approach possesses general applicability in enhancing stability across different models.

A comparison of hallucination control performance between the domain model and the base model further verifies the improvement achieved by the proposed approach. After domain-specific fine-tuning, Qwen-BIM exhibited better performance in managing hallucinations—its total hallucination count decreased from 65 to 44, representing an overall reduction of approximately 32.3%. This finding suggests that the synergistic effect of domain fine-tuning and hallucination control jointly enhances the stability and reliability of defect identification outputs.

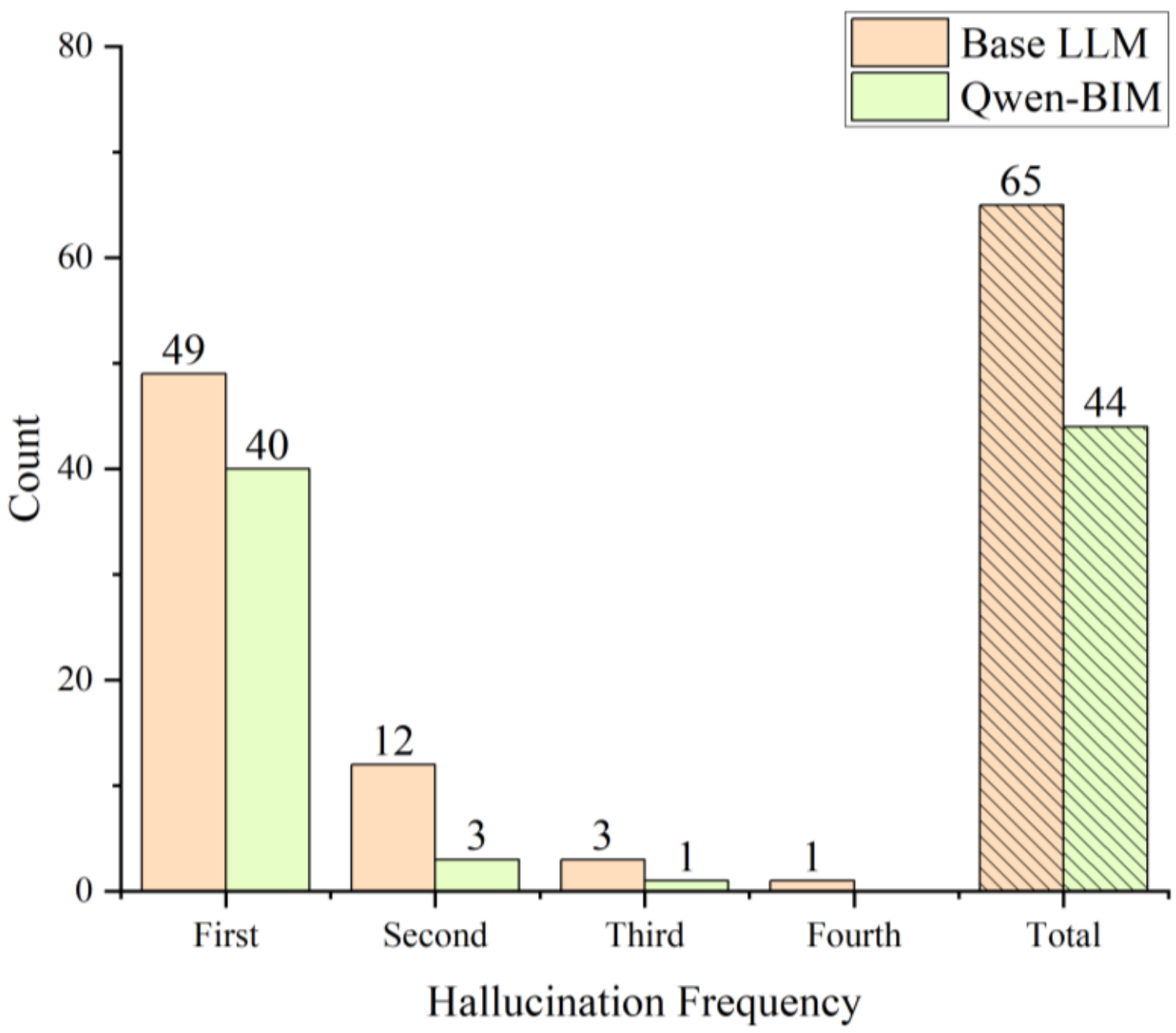

**Fig. 20.** Frequency statistics of hallucination occurrences in defect identification

**Fig. 21** presents the impact of the hallucination control method on defect identification accuracy. It can be observed that the proposed method improves accuracy for both the base model and Qwen-BIM. The base model without any enhancement achieved an identification accuracy of 64%, whereas after fine-tuning and applying the hallucination control method, the accuracy increased to 85%, demonstrating a substantial improvement.

This result indicates that fine-tuning and hallucination control exhibit a synergistic effect: fine-tuning enhances the model's understanding of professional semantics and design logic, while the hallucination control method effectively suppresses erroneous outputs during the generation process. Together, they improve the model's stability and accuracy at both the semantic comprehension and result generation levels.

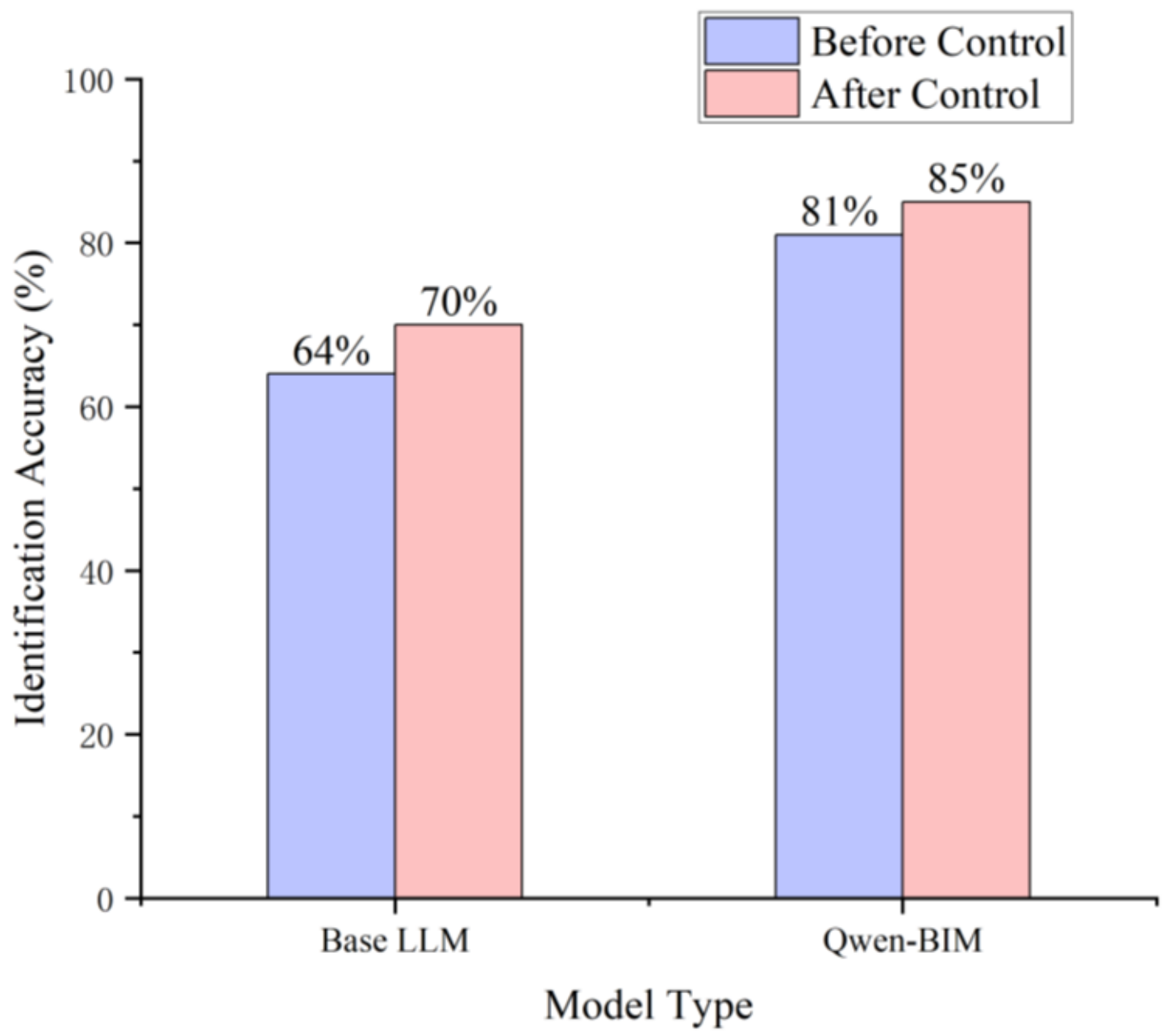

**Fig. 21.** Comparison of identification accuracy before and after applying the hallucination control method

### 4.4 Discussion

While this study successfully applies a domain-specific LLM, a more critical advancement lies in the architectural design of the end-to-end integrated framework. The ablation studies indicate that even when utilizing the base LLM, the framework achieved a 70% identification accuracy and, with the integration of RAG, a 92% rationality rate in generating repair suggestions. The incorporation of the domain-tuned Qwen-BIM further elevated the identification accuracy to 85% and the repair rationality to 94%. This comparative performance highlights that the fundamental effectiveness of the defect identification and repair process is driven by the robust architecture of the framework itself, with the domain-specific model serving as a further performance enhancer.

Furthermore, the integrated framework expands the capability boundaries of automated checking by enabling the identification of rationality defects. By leveraging semantic understanding, the model successfully identifies practical anomalies from normative value ranges, such as a mathematically compliant but practically absurd 30-hour fire resistance limit. This capability effectively bridges the gap between explicit regulatory codes and implicit engineering common sense. However, deploying generative models in strict engineering contexts requires mitigating their inherent hallucination risks. To address this, the proposed hallucination control strategy introduces a deterministic post-validation mechanism. By enforcing key-identifier cross-checks and token-length constraints, the text generation is strictly anchored to the actual BIM components rather than fabricated data. The ability to eliminate 92.5% of hallucinations in a single intervention significantly enhances output stability, demonstrating a highly viable pathway to meet the stringent reliability demands of architectural design workflows.

Despite these advancements, the empirical validation of the current framework is inherently scoped to a single residential building model and specific fire protection codes. However, the core mechanisms of the proposed framework exhibit strong methodological transferability. The component-balanced BIM-to-Text chunking and the semantic retrieval logic within the RAG module are largely agnostic to specific building typologies or regulatory domains. By updating the vector knowledge base and refining the prompt templates, the framework can be readily extended to other complex scenarios, such as commercial high-rises or MEP coordination. Recognizing these empirical boundaries while confirming the framework's structural adaptability provides a robust and transparent foundation for future multidisciplinary applications.

## 5 Conclusion

Human-induced design defects in BIM are inevitable, while current detection and repair remain largely manual and inefficient. Although automation has been explored, existing methods lack generality, and research on automated defect repair is still limited. To address these gaps, this study proposes an integrated framework for multi-category BIM design defect identification and repair suggestion generation. The main research contents and findings are summarized as follows:

(1) This study establishes an end-to-end prototype based on an integrated framework, successfully closing the workflow from raw BIM data input, through defect identification, to the final generation of repair suggestions. Specifically, the framework achieves the identification of multi-category design defects by combining a BIM-to-Text transformation method and prompt learning with the domain-specific LLM. Furthermore, it utilizes RAG coupled with prompt learning to guide the model in producing code-compliant repair suggestions for the identified issues. Finally, a domain-adapted hallucination control mechanism is incorporated into the pipeline to ensure the high precision and engineering reliability of the final outputs.

(2) This study proposes a multi-category defect identification method based on prompt learning and conducts comparative experiments between this method and the traditional rule checking approach in the context of fire code compliance checking for residential BIM models. The results demonstrate that the proposed method achieves a capability expansion. The overall identification accuracy improves from 70% to 85% primarily because the model can automatically address rationality defects. For these rationality issues, the model demonstrates the ability to identify gross deviations from normative value ranges, a task for which purely syntactic rule checking is ill-suited, thereby effectively complementing traditional methods.

(3) To enhance the precision and reliability of defect repair, this study developed a RAG-based defect repair method supported by a vectorized knowledge base comprising 221 data nodes derived from design codes. Experimental validation in typical application scenarios demonstrated that integrating RAG effectively strengthened the model’s normative grounding and external knowledge utilization, enabling the domain-specific model Qwen-BIM to generate more accurate and contextually appropriate repair suggestions, with the rationality rate improving from 79% to 94%, significantly outperforming the base model.

(4) To address the hallucination issues that may arise during the defect identification process

of LLMs, this study proposes two hallucination control methods based on key-identifier validation and token-length thresholds. Experimental results show that while the base model achieved an identification accuracy of 64%, the fine-tuned Qwen-BIM, after applying the proposed control methods, improved the accuracy to 85%, with 92.5% of hallucinations eliminated through a single control intervention. By effectively mitigating fabricated information and excessive generation, this mechanism significantly enhances the model's reliability, thereby providing a practical foundation for integrating LLMs into rigorous engineering workflows.

Ultimately, this study systematically integrates LLMs, prompt learning, RAG, and post-hoc hallucination control to construct an end-to-end prototype. Through this comprehensive pipeline, we have developed and demonstrated a novel, reliable, and reproducible architectural paradigm for integrating generative AI into rigorous engineering workflows, directly addressing key challenges of verification and adaptability. This provides a robust structural foundation that future research can utilize and build upon.

Building upon the proposed integrated framework, future research can be expanded in two main directions. First, since the present study primarily validated the proposed methods in limited scenarios, future work should focus on applying and evaluating the prototype in complex and diverse engineering projects. This includes assessing its accuracy and hallucination rates across various novel defect types, developing more effective strategies to further mitigate hallucinations, and exploring automated defect repair from a multi-disciplinary perspective to synthesize diverse engineering factors. These continuous expansions will ultimately ensure the practicality and reliability of this generative technology in actual rigorous engineering contexts. Second, the integration of LLM-based intelligent agents could be explored to enable real-time interaction between the domain model and native BIM software. This would allow the system not only to identify defects and propose repair suggestions but also to automatically execute model modifications, transitioning towards an intelligent collaborative paradigm encompassing identification, recommendation, and execution.

## CRediT authorship contribution statement

**Jia-Rui Lin:** Conceptualization, Supervision, Methodology, Writing – review & editing, Funding acquisition, Project administration. **Yun-Hong Cai:** Writing – original draft, Writing – review & editing, Methodology, Data curation, Formal analysis, Visualization. **Xiang-Rui Ni:** Writing – original draft, Methodology, Validation, Investigation, Data curation. **Peng Pan:** Conceptualization, Supervision, Project administration, Funding acquisition, Writing – review & editing.

## Data availability

Data will be made available on request.

## Acknowledgments

The authors are grateful for the financial support received from the National Key R&D

Program of China (No. 2023YFC3804600) and the National Natural Science Foundation of China (No. 52378306).

## Appendix A

### A.1 Defect Identification Prompt Template

[System Prompts]

*"You are a professional architectural designer. In the following dialogue, after performing stepwise calculation and reasoning to obtain a result, please provide the final answer in the following format. Regardless of whether any reasoning steps are shown, you must add the string '[Final Answer]:' on a new line, and then, starting from the next line, give your answer to the question in complete sentences, without including any additional information or reasoning process. Please strictly follow the format shown in the example provided after the question when giving the final answer."*

[Rule-Injection Prompts]

*"Please examine each component in the building model according to the following rules to determine whether it meets the requirements. The specific rules are expressed as in the example below:*

*For load-bearing wall components with Fire Resistance Ratings (I, II, III, IV), the combustibility shall be (non-combustible, non-combustible, non-combustible, flame-retardant), and the fire resistance limit shall not be less than (3.00h, 2.50h, 2.00h, 0.50h). This rule actually represents four separate conditions: for load-bearing walls with Fire Resistance Ratings I, II, III, and IV, their combustibility shall be non-combustible, non-combustible, non-combustible, and flame-retardant respectively, and their fire resistance limits shall not be less than 3.00h, 2.50h, 2.00h, and 0.50h, respectively.*

*The complete set of rules is as follows:*

*(1) For load-bearing wall components with Fire Resistance Ratings (I, II, III, IV), combustibility shall be (non-combustible, non-combustible, non-combustible, flame-retardant), and the fire resistance limit shall not be less than (3.00h, 2.50h, 2.00h, 0.50h).*

*(2) For non-load-bearing exterior wall components with Fire Resistance Ratings (I, II, III, IV), combustibility shall be (non-combustible, non-combustible, non-combustible, combustible), and the fire resistance limit shall not be less than (1.00h, 1.00h, 0.50h, no requirement).*

*(3) For non-load-bearing interior wall components with Fire Resistance Ratings (I, II, III, IV), combustibility shall be (non-combustible, non-combustible, flame-retardant, flame-retardant), and the fire resistance limit shall not be less than (0.75h, 0.50h, 0.50h, 0.25h).*

*(4) For floor slab components with Fire Resistance Ratings (I, II, III, IV), combustibility shall be (non-combustible, non-combustible, non-combustible, combustible), and the fire resistance limit shall not be less than (1.50h, 1.00h, 0.50h, no requirement).*

*(5) For evacuation stair components with Fire Resistance Ratings (I, II, III, IV), combustibility*

*shall be (non-combustible, non-combustible, non-combustible, combustible), and the fire resistance limit shall not be less than (1.50h, 1.00h, 0.50h, no requirement).*

*(6) For accessible flat roof components with Fire Resistance Ratings (I, II), combustibility shall be (non-combustible, non-combustible), and the fire resistance limit shall not be less than (1.50h, 1.00h).*

*Based on the six rules above, please check whether each component in the building model complies with the specified requirements. The inspection scope is limited to the given components. If any missing values or abnormal attribute values are found, please identify them as well. If no problems are detected, leave the section following [Final Answer] blank."*

[Few-Shot Prompts]

*"After [Final Answer], only list the ID of each problematic component and its specific issue. Separate the ID and the problem description with a comma, and separate different components with semicolons. For example: 348253, fire performance is combustible, not compliant; 350236, fire performance missing, not compliant."*

**A.2 Repair Suggestion Generation Prompt Template**

[Textualized Model Data]

*"This is a part of a building model, which includes wall components.*

*The following provides a description of the wall components, specifying their name, ID, floor location, geometric parameters including thickness and height, functional properties such as load-bearing status and interior or exterior type, and fire safety attributes including fire resistance rating, combustibility, and fire resistance limit.*

*The name is 200 reinforced concrete exterior wall, ID is 348253, located on 1F; wall thickness is 200 mm, height is 3050 mm, it is a non-load-bearing wall, and it is an exterior wall; its fire resistance rating is Grade II, combustibility is combustible, and the fire resistance limit is 4 hours.*

*The name is 200 aerated concrete block wall, ID is 350191, located on 1F; wall thickness is 200 mm, height is 2900 mm, it is a non-load-bearing wall, and it is an interior wall; its fire resistance rating is Grade II, combustibility is non-combustible, and the fire resistance limit is 4 hours.*

*The name is 200 aerated concrete block wall, ID is 350236, located on 1F; wall thickness is 200 mm, height is 2900 mm, it is a non-load-bearing wall, and it is an interior wall; its fire resistance rating is Grade II, combustibility information is missing, and the fire resistance limit is 4 hours.*

*The name is 200 aerated concrete block wall, ID is 350606, located on 1F; wall thickness is 200 mm, height is 2900 mm, it is a non-load-bearing wall, and it is an exterior wall; its fire resistance rating is Grade II, combustibility is non-combustible, and the fire resistance limit is 4 hours.*

*The name is 200 aerated concrete block wall, ID is 350758, located on 1F; wall thickness is 200 mm, height is 3050 mm, it is a non-load-bearing wall, and it is an exterior wall; its fire resistance rating is Grade II, combustibility is Class A non-combustible, and the fire resistance limit*

*is 4 hours.*

*The name is 200 aerated concrete block wall, ID is 350838, located on 1F; wall thickness is 200 mm, height is 3050 mm, it is a non-load-bearing wall, and it is an exterior wall; its fire resistance rating is Grade II, combustibility is non-combustible, and the fire resistance limit is 0.3 hours.*

*The name is 200 aerated concrete block wall, ID is 350925, located on 1F; wall thickness is 200 mm, height is 3050 mm, it is a non-load-bearing wall, and it is an exterior wall; its fire resistance rating is Grade II, combustibility is non-combustible, and the fire resistance limit is 0.3 hours.*

*The name is 200 aerated concrete block wall, ID is 351144, located on 1F; wall thickness is 200 mm, height is 2900 mm, it is a non-load-bearing wall, and it is an interior wall; its fire resistance rating is Grade II, combustibility is Class A non-combustible, and the fire resistance limit is 4 hours.*

*The name is 200 aerated concrete block wall, ID is 351364, located on 1F; wall thickness is 200 mm, height is 2900 mm, it is a non-load-bearing wall, and it is an interior wall; its fire resistance rating is Grade II, combustibility is non-combustible, and the fire resistance limit is 4 hours.*

*The name is 100 aerated concrete block wall, ID is 351602, located on 1F; wall thickness is 100 mm, height is 2900 mm, it is a non-load-bearing wall, and it is an interior wall; its fire resistance rating is Grade II, combustibility is combustible, and the fire resistance limit is 3.75 hours."*

[System Prompts]

*"You are a professional architectural designer. In the following dialogue, after performing stepwise calculation and reasoning to obtain a result, please provide the final answer in the following format. Regardless of whether any reasoning steps are shown, you must add the string '[Final Answer]:' on a new line, and then, starting from the next line, give your answer to the question in complete sentences, without including any additional information or reasoning process. Please strictly follow the format shown in the example provided after the question when giving the final answer."*

[Defect Repair Prompt]

*"There are some defects in the components within this part of the building model. The component IDs and their specific issues are as follows:*

*(Example input for a specific inference task):*

*348253, combustibility is combustible, not compliant; 350236, combustibility information is missing, not compliant; 350838, the fire resistance limit is 0.3 hours, not compliant; 350925, the fire resistance limit is 0.3 hours, not compliant.*

*Please help me repair these defects so that the components comply with the requirements."*

[Few-Shot Prompts]

*"After [Final Answer], only list the ID of each problematic component and its specific issue.*

*Separate the ID and the problem description with a comma, and separate different components with semicolons. For example: 348253, fire performance is combustible, not compliant; 350236, fire performance missing, not compliant."*

[RAG-retrieved knowledge]

*"(Example of rules retrieved by the RAG module):*

*The combustibility of non-load-bearing exterior wall components with a Grade II fire resistance rating is non-combustible, and the fire resistance limit is no less than 1.00 hour.*

*The combustibility of non-load-bearing interior wall components with a Grade II fire resistance rating is non-combustible, and the fire resistance limit is no less than 0.50 hours.*

*The combustibility of load-bearing wall components with a Grade II fire resistance rating is non-combustible, and the fire resistance limit is no less than 2.50 hours."*

[RAG prompt]

*"The following content consists of code provisions retrieved from the knowledge base based on the user's query. Please first determine whether these contents are relevant to the question, and then answer the user's question based on the relevant provisions."*